\documentclass{ieeeaccess}

\usepackage{cite}
\usepackage{amsmath,amssymb,amsfonts}
\usepackage{algorithmic}
\usepackage{graphicx}
\usepackage{textcomp}
\usepackage{bm}
\usepackage{lineno}
\usepackage{float}         
\usepackage{multirow}      
\usepackage{booktabs}      
\usepackage{siunitx}       
\usepackage{soul}          
\usepackage{url}
\usepackage{hyperref}
\hypersetup{
    colorlinks=true,   
    linkcolor=blue,    
    citecolor=blue,    
    urlcolor=blue      
}

\definecolor{darkyellow}{rgb}{0.8, 0.6, 0}
\definecolor{mycolor}{rgb}{0.1,0.2,0.3}

\sethlcolor{darkyellow}

\renewcommand{\arraystretch}{1.5}
\renewcommand{\thetable}{\Roman{table}}

\makeatletter
\AtBeginDocument{
    \DeclareMathVersion{bold}
    \SetSymbolFont{operators}{bold}{T1}{times}{b}{n}
    \SetSymbolFont{NewLetters}{bold}{T1}{times}{b}{it}
    \SetMathAlphabet{\mathrm}{bold}{T1}{times}{b}{n}
    \SetMathAlphabet{\mathit}{bold}{T1}{times}{b}{it}
    \SetMathAlphabet{\mathbf}{bold}{T1}{times}{b}{n}
    \SetMathAlphabet{\mathtt}{bold}{OT1}{pcr}{b}{n}
    \SetSymbolFont{symbols}{bold}{OMS}{cmsy}{b}{n}
    \renewcommand\boldmath{\@nomath\boldmath\mathversion{bold}}
}
\makeatother

\def\BibTeX{{\rm B\kern-.05em{\sc i\kern-.025em b}\kern-.08em
    T\kern-.1667em\lower.7ex\hbox{E}\kern-.125emX}}

\begin{document}
\history{Date of publication xxxx 00, 0000, date of current version xxxx 00, 0000.}
\doi{10.1109/ACCESS.2024.0429000}

\title{A Benchmark Dataset and a Framework for Urdu Multimodal Named Entity Recognition}
\author{\uppercase{Hussain Ahmad}\authorrefmark{1},
\uppercase{Qingyang Zeng}\authorrefmark{1},
\uppercase{Jing Wan}\authorrefmark{1}}

\address[1]{College of Information Science and Technology, Beijing University of Chemical Technology, Beijing, 100029, China}

\tfootnote{}

\markboth
{Hussain Ahmad \headeretal: A Benchmark Dataset and a Framework for Urdu Multimodal Named Entity Recognition}
{Hussain Ahmad \headeretal: A Benchmark Dataset and a Framework for Urdu Multimodal Named Entity Recognition}

\corresp{Corresponding author: Jing Wan (e-mail: wanj@mail.buct.edu.cn).}

\begin{abstract}
The emergence of multimodal content, particularly text and images on social media, has positioned Multimodal Named Entity Recognition (MNER) as an increasingly important area of research within Natural Language Processing. Despite progress in high-resource languages such as English, MNER remains underexplored for low-resource languages like Urdu. The primary challenges include the scarcity of annotated multimodal datasets and the lack of standardized baselines. To address these challenges, we introduce the U-MNER framework and release the Twitter2015-Urdu dataset, a pioneering resource for Urdu MNER. Adapted from the widely used Twitter2015 dataset, it is annotated with Urdu-specific grammar rules. We establish benchmark baselines by evaluating both text-based and multimodal models on this dataset, providing comparative analyses to support future research on Urdu MNER. The U-MNER framework integrates textual and visual context using Urdu-BERT for text embeddings and ResNet for visual feature extraction, with a Cross-Modal Fusion Module to align and fuse information. Our model achieves state-of-the-art performance on the Twitter2015-Urdu dataset, laying the groundwork for further MNER research in low-resource languages.
\end{abstract}

\begin{keywords}
Multimodal Named Entity Recognition, Urdu, Social Media, Urdu Multimodal NER Dataset, Cross-Modal Attention
\end{keywords}

\titlepgskip=-21pt

\maketitle

\section{Introduction}
\label{sec:introduction}
Multimodal Named Entity Recognition (MNER) is a task in Natural Language Processing that identifies and classifies named entities from text into predefined categories, such as people, organizations, and locations, with the assistance of images to enhance classification accuracy~\cite{mai2024dynamic, yu2020improving}. Unlike standard Named Entity Recognition (NER), which relies solely on text to detect entities, MNER incorporates visual content to resolve ambiguities and provide additional context for more precise identification. This is especially valuable in domains like social media, where text is often accompanied by images that help clarify the meaning of entities, leading to improved recognition performance~\cite{liu2024dghc}. 

\begin{figure}[ht] 
    \centering
    \includegraphics[width=0.75\columnwidth]{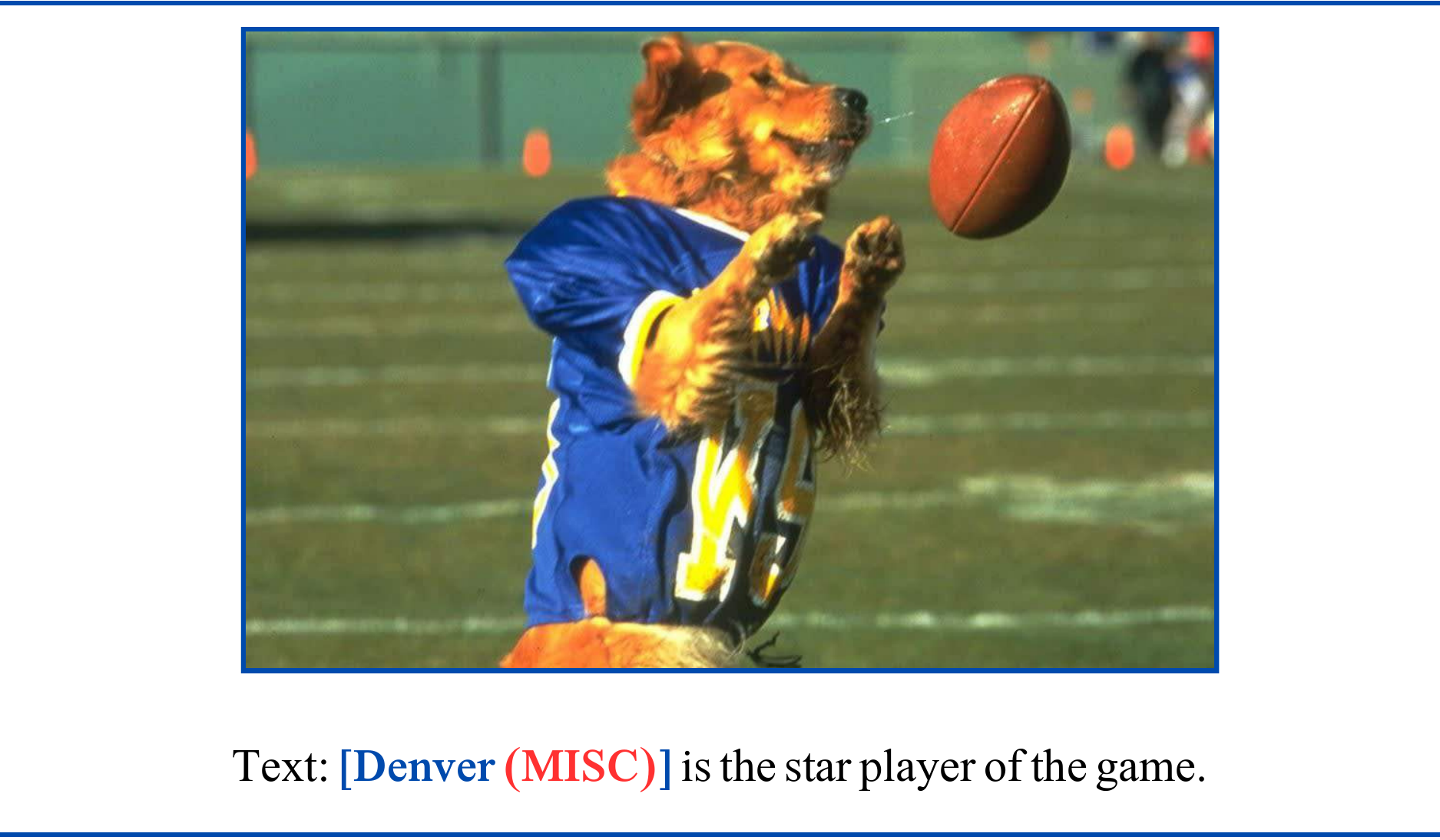} 
    \caption{An example of Multimodal Named Entity Recognition, with the named entity and its type highlighted in brackets.}
    \label{fig0}
\end{figure}

Visual information resolves ambiguities where textual context is insufficient for accurate entity classification. For example, the text "Denver is the star player of the game" alone leaves it unclear whether "Denver" refers to a person or something else. However, the accompanying image (as shown in Figure~\ref{fig0}) clarifies that "Denver" is a dog, correctly classified as miscellaneous (MISC). This demonstrates how visual context enhances textual data, improving the precision of named entity recognition in challenging scenarios.

While considerable advancements have been achieved in MNER for high-resource languages (e.g., English), its application to low-resource languages, such as Urdu, remains significantly underexplored~\cite{liang2024consistency}. Urdu is the native language of 70 million people and the second language of over 100 million speakers across Pakistan and India. A critical challenge lies in the limited availability of annotated datasets and the absence of well-defined baselines for Urdu MNER, significantly hindering model development and evaluation for this linguistically rich language. These challenges underscore the necessity of a dedicated dataset and a tailored framework for Urdu~\cite{aliyu2024sentiment}. Bridging this gap is crucial for advancing MNER research in low-resource languages and broadening its applicability across diverse linguistic contexts.

To address this gap, we introduce the Twitter2015-Urdu dataset, a pioneering MNER dataset specifically designed for Urdu. This dataset is constructed by translating and annotating the widely-used Twitter2015 English dataset into Urdu, ensuring that both the text and corresponding images are culturally and linguistically relevant. Rather than relying on simple translations, the Twitter2015-Urdu dataset adapts the original dataset to account for language-specific features of Urdu, such as its rich morphology, complex script, and context-sensitive expressions, providing a valuable benchmark for developing and evaluating MNER systems in Urdu. 

To further enhance the utility of the Twitter2015-Urdu dataset, we conducted extensive experiments to establish performance baselines for Urdu MNER. We evaluated both text-based and multimodal models, enabling a comprehensive assessment of performance under diverse conditions. The resulting benchmarks serve as a key reference for future studies, enabling comparative analysis and supporting the development of Urdu MNER systems in the research community.

We propose U-MNER, a novel framework for Urdu MNER that leverages both textual and visual information to enhance entity recognition performance specifically within social media contexts. Our framework utilizes Urdu-BERT~\cite{ahmed2024urduNER} for contextualized word embeddings, which captures the linguistic nuances of Urdu, and Residual Network (ResNet) for visual feature extraction, chosen for its ability to preserve and integrate complex visual data. These components are integrated through cross-modal attention mechanisms, allowing the model to effectively disambiguate entities and improve classification accuracy by leveraging the complementary nature of text and image data. Extensive experiments reveal that the proposed model achieves state-of-the-art performance on the Twitter2015-Urdu dataset, outperforming existing methods. By incorporating multimodal data, our model addresses specific challenges in Urdu, such as ambiguous entity types and spelling variations, underscoring the potential of our approach to advance MNER research for low-resource languages.

The key contributions of this paper are summarized as follows:
\begin{itemize} 
   \item We introduce the Twitter2015-Urdu dataset, the first dataset for Urdu MNER, adapted from the Twitter2015 English dataset and annotated with Urdu-specific linguistic features. This dataset serves as a foundational resource for advancing MNER research in Urdu and similar low-resource languages. 
   
    \item We establish comprehensive baselines for Urdu MNER by benchmarking text-based and multimodal models on the Twitter2015-Urdu dataset. This effort facilitates comparative analysis and provides a strong foundation for future Urdu MNER studies.

    \item We present a novel framework U-MNER that integrates textual and visual data through cross-modal attention for Urdu MNER. Our model achieves state-of-the-art performance on the Twitter2015-Urdu dataset. Ablation studies further confirm the impact of our proposed modules on model effectiveness.
\end{itemize}

\section{RELATED WORK}
\subsection{Multimodal Named Entity Recognition}
MNER enhances traditional NER by integrating both textual and visual data, improving entity recognition in contexts where images offer additional contextual cues, such as on social media~\cite{lu2018visual}. By combining modalities, MNER can resolve ambiguities in text-only data. 

Early MNER models utilized visual features to supplement text-based NER~\cite{wang2024boosting,bao2023mpmrc}, but these approaches struggled with irrelevant visual content that could mislead the recognition process. More recent advancements, such as the unified multimodal transformer~\cite{yu2020improving}, introduced a visual gate to dynamically filter out irrelevant visual information, aligning images with relevant textual entities. This refinement has significantly improved performance by allowing models to focus solely on pertinent visual data, mitigating the impact of noisy or unrelated images. Further advancements in MNER include models like RpBERT~\cite{sun2021rpbert} and ITA~\cite{wang2022ita}, which leverage cross-modal attention and object alignment to improve the interaction between text and image data. Other studies~\cite{jia2023mnerqg},~\cite{xu2022maf} have introduced tasks like query grounding and text-image relation classification to strengthen modality integration and aid in resolving complex ambiguities. In the context of low-resource languages, Yamini et al. \cite{yamini2024kurdsm} introduced KurdSM, a transformer-based model designed for abstractive text summarization. This approach underscores the challenges associated with processing languages with limited resources and the need for specialized corpora. It highlights how transformer-based models can be adapted to such tasks, aligning with the focus of this work on multimodal named entity recognition for Urdu, a linguistically complex low-resource language.

While these improvements have advanced MNER in high-resource languages, they are challenging to adapt to low-resource languages like Urdu due to limited annotated multimodal data and Urdu's linguistic complexity, necessitating novel approaches and resources to bridge this gap.

\subsection{Challenges in Urdu Named Entity Recognition}
Urdu presents several unique challenges for Named Entity Recognition (NER), primarily due to its linguistic complexity, morphological richness, and the scarcity of dedicated resources. These challenges are outlined below:

\textbf{Absence of Capitalization:}
In English, capitalization serves as a primary cue for identifying named entities. However, Urdu lacks this orthographic feature, which makes it more challenging to distinguish named entities from common words, as the language does not provide such visual markers.

\textbf{Context Sensitivity:}
The classification of a named entity token is highly context-dependent, with its classification potentially changing based on the surrounding text. A token that represents one type of named entity in a given context may be classified differently or may not be considered an NE at all in another context. For instance, in the sentence \textquotedblleft\raisebox{-0.2\height}{\includegraphics[height=1.2em]{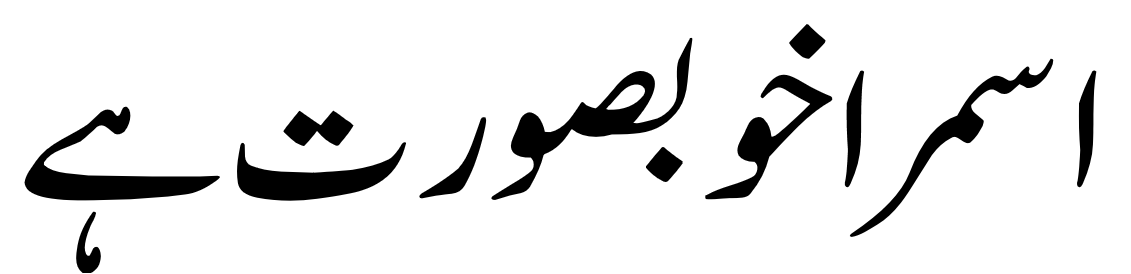}}\textquotedblright\ (Asmara is beautiful), the token \textquotedblleft\raisebox{-0.2\height}{\includegraphics[height=1.0em]{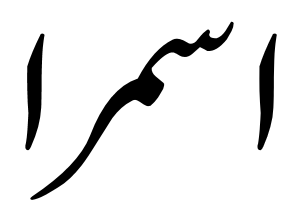}}\textquotedblright\ (Asmara) refers to a person. However, in the sentence \textquotedblleft\raisebox{-0.2\height}{\includegraphics[height=1.5em]{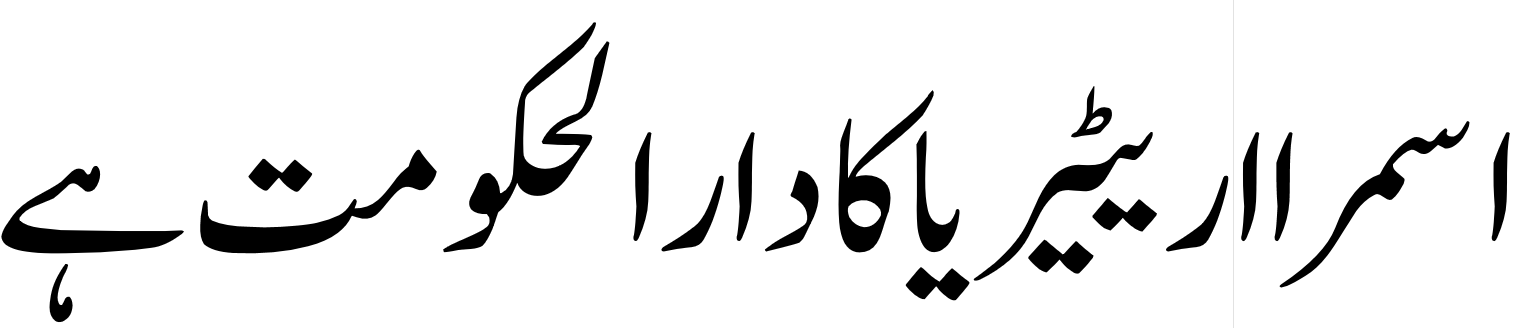}}\textquotedblright\ (Asmara is the capital of Eritrea), the same token refers to a location. This context sensitivity poses challenges in named entity recognition, requiring models to effectively disambiguate entity types based on surrounding linguistic cues.

\textbf{Inflectional and Agglutinative Nature:}
Urdu exhibits a highly inflectional and agglutinative morphology, wherein words often undergo structural transformations when combined with others~\cite{kanwal2019}. These modifications can alter or eliminate their status as named entities, thereby complicating the entity recognition process. For example, \textquotedblleft\raisebox{-0.2\height}{\includegraphics[height=1.0em]{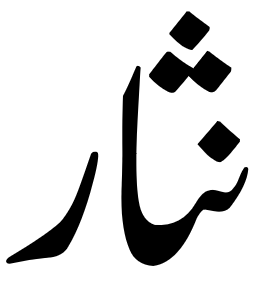}}\textquotedblright\ (Nisar) is frequently used as a proper noun to denote a person's name. However, when interpreted in its literal, compositional form—combining \textquotedblleft\raisebox{-0.2\height}{\includegraphics[height=1.0em]{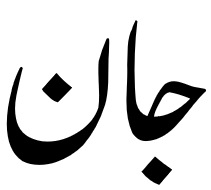}}\textquotedblright\ (life) and \textquotedblleft\raisebox{-0.2\height}{\includegraphics[height=1.0em]{images/urdutextimages/Nisar.png}}\textquotedblright\ (devoted)—it conveys the meaning "self-sacrificing" or "devoted" and no longer functions as a named entity. Such morphological ambiguity introduces challenges in accurately determining entity boundaries and classifying tokens, particularly in automated systems that rely on static lexical cues.

\textbf{Free Word Order:}
Urdu allows for a flexible word order, which means that the syntactic structure of a sentence can be altered without changing its semantic meaning. For instance, both \textquotedblleft\raisebox{-0.2\height}{\includegraphics[height=1.4em]{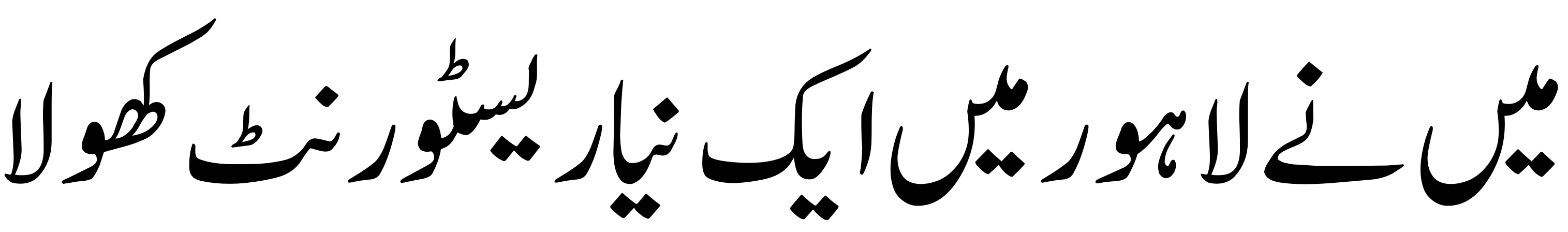}}\textquotedblright\ (I opened a new restaurant in Lahore) and \textquotedblleft\raisebox{-0.2\height}{\includegraphics[height=1.4em]{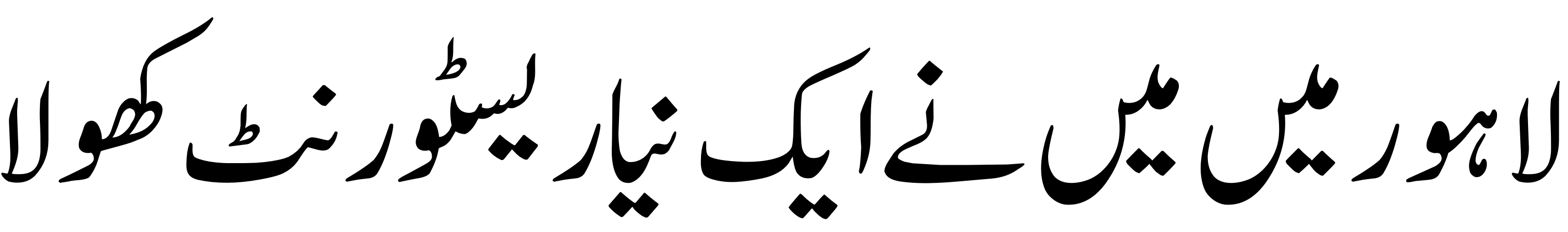}}\textquotedblright\ (In Lahore, I opened a new restaurant) convey the same meaning, despite the rearrangement of words. This flexibility presents challenges for sequence-based models, which typically rely on fixed syntactic structures for entity recognition.

\textbf{Spelling Variations and Orthographic Ambiguity:}
Spelling inconsistencies and orthographic variations are common in Urdu, particularly in informal or social media texts. A single named entity may appear in multiple forms, such as \textquotedblleft\raisebox{-0.2\height}{\includegraphics[height=0.7em]{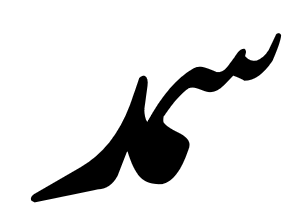}}\textquotedblright\, \textquotedblleft\raisebox{-0.2\height}{\includegraphics[height=1.0em]{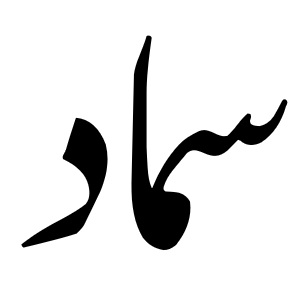}}\textquotedblright\, and \textquotedblleft\raisebox{-0.2\height}{\includegraphics[height=1.0em]{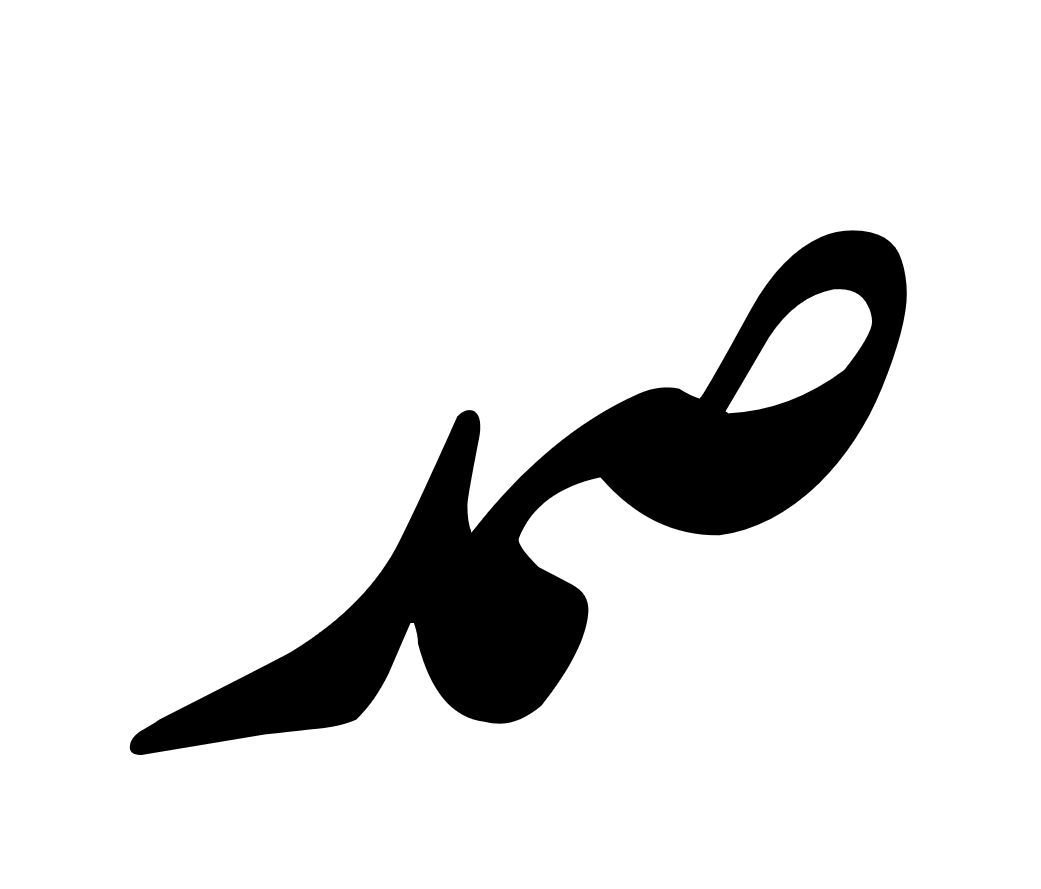}}\textquotedblright\, all representing the name "Samad". Additionally, phrases like \textquotedblleft\raisebox{-0.2\height}{\includegraphics[height=1.2em]{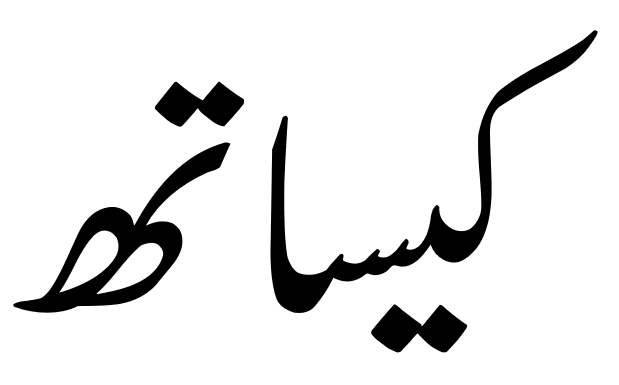}}\textquotedblright\ and \textquotedblleft\raisebox{-0.2\height}{\includegraphics[height=1.1em]{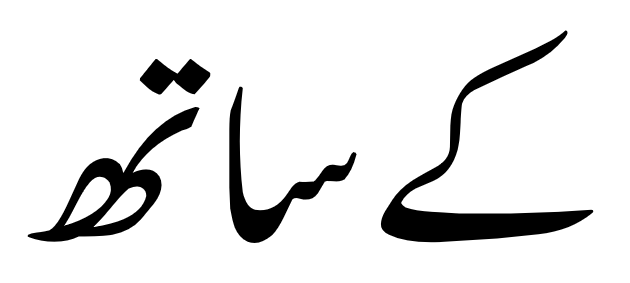}}\textquotedblright\ (both meaning "with") may appear with or without a space, complicating tokenization and negatively affecting entity recognition accuracy.

\subsection{Urdu MNER Dataset}
Despite advancements in MNER for high-resource languages, low-resource languages like Urdu lack equivalent multimodal datasets. Existing resources for Urdu NER task, such as the CRULP Corpus and the UCREL Corpus~\cite{shafi2023semantic}, primarily contain text-only annotations within Urdu news articles, restricting their suitability for multimodal research. Additionally, multilingual datasets such as WikiDiverse~\cite{wang2022wikidiverse} and OntoNotes offer some text-image pairs but are not tailored to the unique morphological, syntactic, and semantic complexities of Urdu, thereby limiting their utility for Urdu-specific MNER research. Twitter2015~\cite{zhang2018coattention} has become a widely adopted benchmark dataset for MNER and has been extensively used in prior studies to evaluate the performance of MNER models. However, it exclusively features English-language content, limiting its applicability to linguistically diverse regions. 

To bridge this gap, we introduce the Twitter2015-Urdu MNER dataset, developed as an Urdu-specific adaptation of the original Twitter2015 English dataset. This dataset is meticulously adapted with culturally grounded annotations to guarantee the faithful and context-specific representation of entity types in Urdu. With a balanced distribution across training, validation, and test sets, the Twitter2015-Urdu dataset is designed to support rigorous experimentation in Urdu MNER, making it a vital resource for advancing multimodal research in low-resource languages.
\begin{figure*}[ht!]
    \centering
    \includegraphics[width=0.8\textwidth]{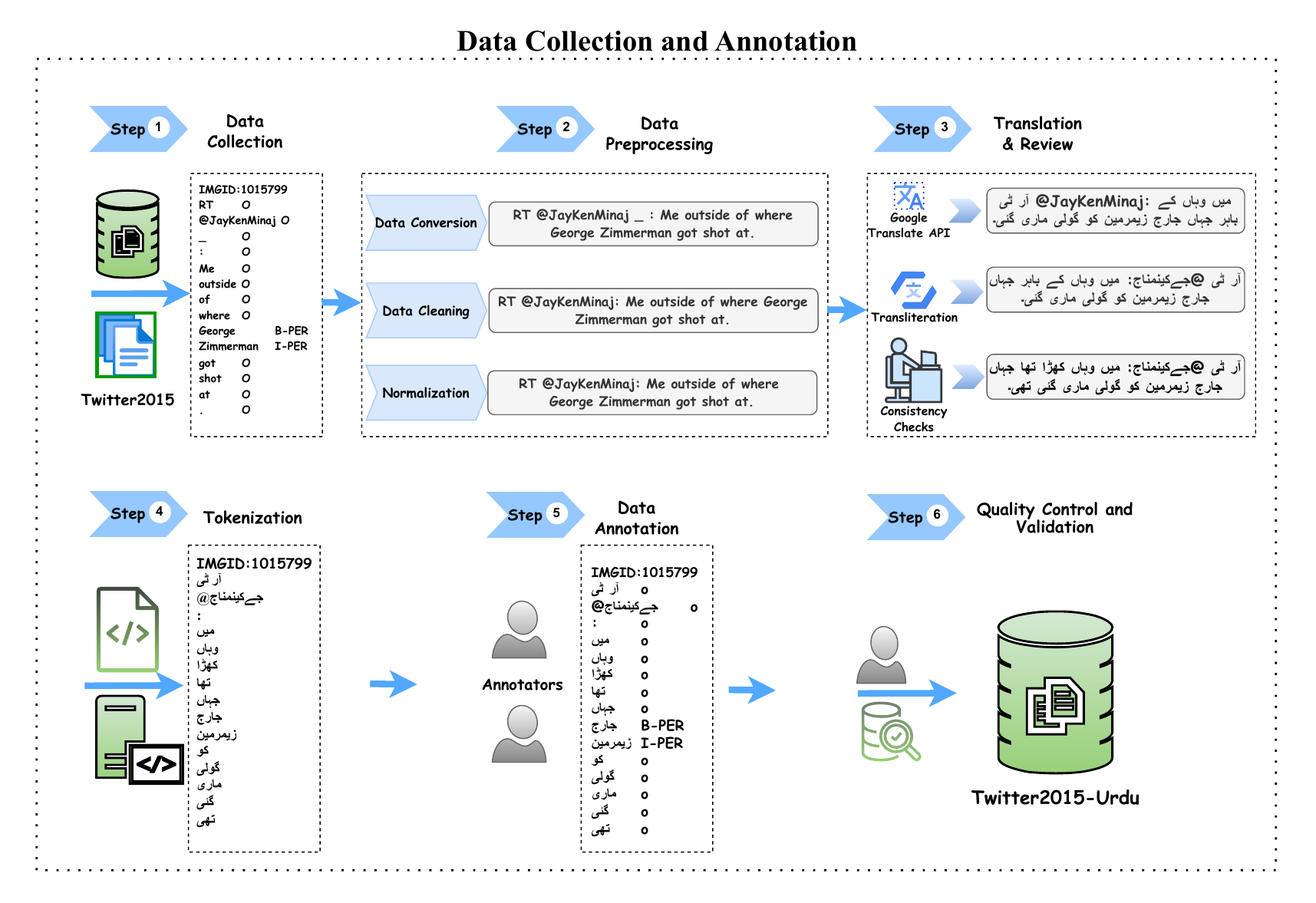}
    \caption{Dataset development pipeline.}
    \label{fig1}
    \vspace{-0.1in}  
\end{figure*}
\section{CORPUS CREATION}
To address the scarcity of publicly available datasets for Urdu MNER, we introduce the Twitter2015-Urdu dataset, which is a version of the Twitter2015 dataset adapted to the Urdu language. This adaptation ensures the dataset reflects the linguistic and cultural characteristics of Urdu, making it suitable for low-resource language research. 

The process of creating this corpus involved several critical stages: data collection, data preprocessing, translation and review, tokenization, data annotation, and quality control and validation, as illustrated in Figure~\ref{fig1}. Each of these steps was crucial in ensuring the dataset’s relevance and applicability to Urdu MNER tasks. This chapter provides a detailed account of each stage, highlighting the methodologies, challenges, and innovations involved in constructing the Twitter2015-Urdu dataset.

\subsection{Data Collection}
For the creation of the Twitter2015-Urdu dataset, data is collected directly from the Twitter2015 English dataset, which provided structured files containing pre-aligned text-image pairs across training, validation, and test sets. The collection process involves extracting these pairs from each set, preserving the original multimodal format. By sourcing directly from the structured files, the data collection ensures that no additional pairing is needed, allowing for a seamless transition to the preprocessing and adaptation stages for Urdu.

\subsection{Data Preprocessing}
Preprocessing is a crucial step in preparing the dataset for translation and annotation, involving several key tasks designed to enhance the dataset's quality and consistency~\cite{maharana2022review}. These steps ensure that the integrity of the original content is maintained while setting the foundation for high-quality translation and annotation.

\subsubsection{\textbf{Tokenized Data Conversion}} The initial dataset is in a tokenized format, where each word or phrase is associated with its respective named entity label. To facilitate translation, the dataset is converted into plain text by removing these entity labels while preserving sentence structures. This step is vital for maintaining context and meaning, ensuring that the text is coherent and ready for further processing.

\subsubsection{\textbf{Data Cleaning}} Once the data is converted into plain text, a series of cleaning steps are undertaken to ensure a uniform format suitable for translation. This includes removing unnecessary characters such as special symbols, emojis, and non-alphanumeric characters that might not translate well into Urdu, thereby minimizing noise and potential translation errors. Additionally, URLs and user mentions are either anonymized or removed based on their relevance, ensuring that only meaningful content is retained for translation.

\subsubsection{\textbf{Normalization}} Normalization is applied to ensure consistency across the dataset, reducing variability and facilitating accurate translation and annotation. This process involves converting the text to a consistent case, typically lowercase, to eliminate case-related variations that could complicate translation. Redundant whitespaces are removed, and consistent spacing is applied between words and punctuation to improve readability and uniformity. Additionally, punctuation marks are standardized to ensure consistent usage throughout the dataset, enhancing clarity and coherence.

\subsection{Translation \& Review}
The translation of the Twitter2015 dataset from English to Urdu is a critical step in developing a resource that aligns with the linguistic and cultural nuances of Urdu-speaking communities. This process involves multiple stages, each aimed at ensuring that the translated dataset maintained the integrity of the original content while being contextually appropriate for Urdu speakers. Maintaining the original meaning of the English text while translating into Urdu requires substantial manual correction, particularly for idiomatic expressions. Translators must carefully adapt sentence structures to ensure grammatical correctness in Urdu while retaining the original message. An iterative review process ensures high-quality translations, with multiple rounds of review by linguists and native speakers. Contextual adjustments are necessary to ensure the translated text is culturally appropriate and relatable to Urdu speakers. This detailed translation process ensures that the Twitter2015-Urdu dataset maintains high standards of accuracy and relevance, making it a valuable resource for advancing MNER research in low-resource languages.

\subsubsection{\textbf{Automated Translation}} The initial step in the translation process involves utilizing automated translation tools to convert the English text into Urdu. The Google Translate API is employed to provide a preliminary translation. This automated approach is chosen for its efficiency and ability to handle large volumes of text quickly. However, given the syntactic and morphological differences between English and Urdu, automated translation serves as only a foundational step, requiring significant refinement to achieve accuracy and contextual relevance.

\subsubsection{\textbf{Human Review }} Following automated translation, a meticulous review is conducted by native Urdu speakers to ensure the text's accuracy and cultural appropriateness. This review is essential for correcting errors introduced by automated tools and aligning translations with natural Urdu usage. Native speakers assessed linguistic accuracy, focusing on grammatical correctness and coherent sentence structures. They also ensure contextual relevance by adjusting translations to account for cultural nuances and idiomatic expressions. Additionally, efforts are made to maintain consistency in terminology and style across the dataset, ensuring uniformity in translated expressions and phrases. 

\subsubsection{\textbf{Cultural Adaptation and Transliterations}} To enhance the cultural relevance of the translated text, additional modifications are made to align expressions and references with the cultural and contextual nuances of Urdu-speaking audiences. Localization involves adjusting culturally specific terms and references to ensure they were relatable and understandable to Urdu speakers. English idioms that lack direct Urdu equivalents are adapted into culturally relevant expressions. For example, the English idiom \textquotedblleft It's raining cats and dogs\textquotedblright \hspace{0.01cm} is translated into \textquotedblleft\raisebox{-0.2\height}{\includegraphics[height=1.3em]{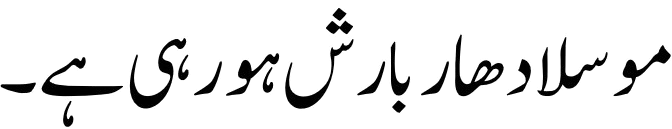}}\textquotedblright\ to convey the same meaning in a culturally appropriate way.

For words or phrases that do not have direct equivalents in Urdu, transliteration is used to maintain the original term's phonetic sound while making it readable in the Urdu script. For instance, certain technological terms, brand names, or specialized jargon often require transliteration to preserve their intended meaning. An example is the term \textquotedblleft smartphone\textquotedblright, which might be transliterated as \textquotedblleft\raisebox{-0.2\height}{\includegraphics[height=1em]{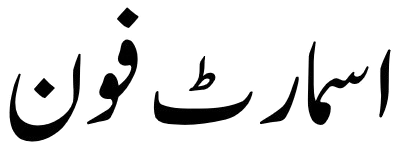}}\textquotedblright\ rather than attempting a literal translation that could lose the word's intended nuance. Transliterations are particularly useful in maintaining clarity and authenticity for specific terms that are globally recognized and do not have a culturally appropriate translation in Urdu.

\subsubsection{\textbf{Consistency Checks}} 
Urdu’s unique features introduce potential inconsistencies. For instance, location names has multiple possible translations, such as \textquotedblleft New York\textquotedblright, which could appear as \textquotedblleft\raisebox{-0.2\height}{\includegraphics[height=1.1em]{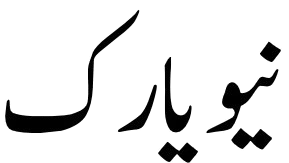}}\textquotedblright , \textquotedblleft\raisebox{-0.2\height}{\includegraphics[height=1em]{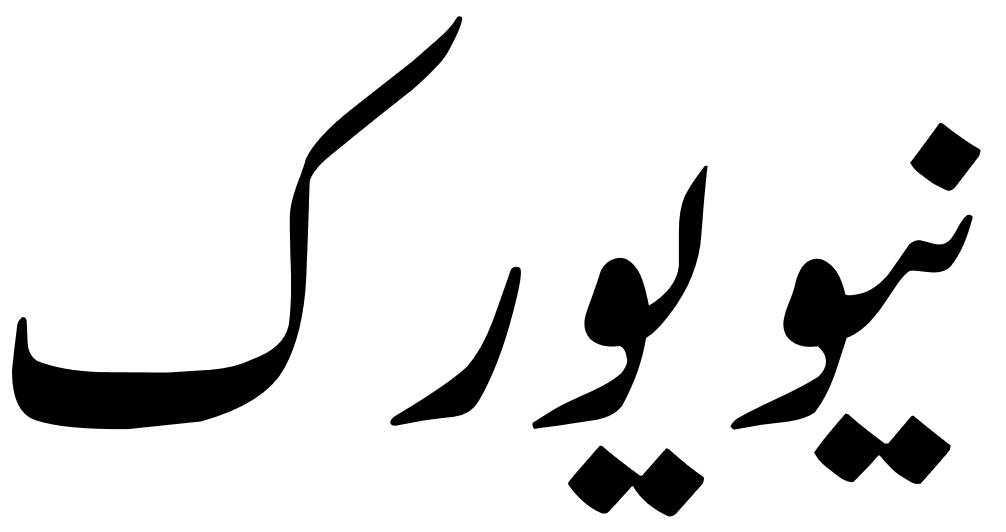}}\textquotedblright , or even\ \textquotedblleft \raisebox{-0.2\height}{\includegraphics[height=1.1em]{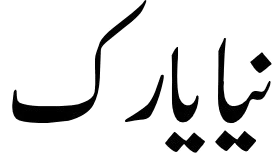}}\textquotedblright\ if not standardized. 
Similarly, terms like \textquotedblleft United Nations\textquotedblright\ has multiple translation options, such as \textquotedblleft\raisebox{-0.2\height}{\includegraphics[height=1.2em]{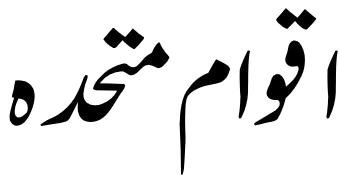}}\textquotedblright\ and \textquotedblleft\raisebox{-0.2\height}{\includegraphics[height=1.1em]{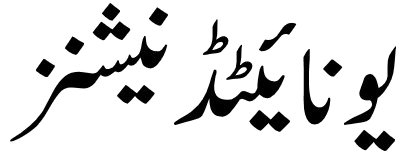}}\textquotedblright, leading to potential variations. Transliteration inconsistencies are also observed for foreign terms like \textquotedblleft Wi-Fi\textquotedblright, which could appear as either \textquotedblleft\raisebox{-0.2\height}{\includegraphics[height=1.1em]{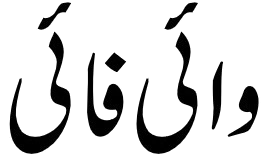}}\textquotedblright\ or \textquotedblleft\raisebox{-0.2\height}{\includegraphics[height=1.2em]{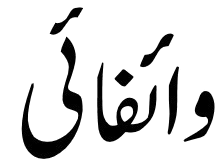}}\textquotedblright. These inconsistencies are addressed through consistent application of predefined transliteration guidelines and manual review processes to ensure clarity and uniformity. By adhering to these guidelines, we maintain consistency across the dataset, preserving its accuracy and reliability for further research and application.

\subsection{Tokenization}
Following translation, we use Urduhack~\cite{zoya2023assessing}, a specialized NLP library for Urdu, to tokenize the text into manageable units, such as words or phrases, for annotation. Designed specifically for Urdu, Urduhack includes tools for normalization, tokenization, and preprocessing, addressing the unique challenges posed by Urdu’s linguistic structure. Tokenization is crucial in Urdu due to its right-to-left, cursive script, which often lacks clear word boundaries, complicating segmentation. Furthermore, as an agglutinative language, Urdu combines prefixes, suffixes, and infixes that alter word forms and meanings within sentences~\cite{ali2017pattern}. For instance, \textquotedblleft\raisebox{-0.2\height}{\includegraphics[height=0.9em]{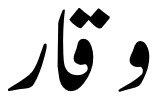}}\textquotedblright\ (Waqar, meaning \textquotedblleft dignity\textquotedblright  \hspace{0.01cm} or \textquotedblleft prestige\textquotedblright) can transform into \textquotedblleft\raisebox{-0.2\height}{\includegraphics[height=1em]{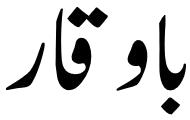}}\textquotedblright\ (BaWaqar, meaning \textquotedblleft dignified\textquotedblright), altering both grammatical role and semantic meaning. Consistent tokenization is necessary to recognize and interpret such morphological changes accurately.
Urduhack performs efficient tokenization of standard Urdu text, but minor errors were observed when handling informal language and compound words, particularly in social media contexts where slang and non-standard spellings are common. For example, Urduhack might struggle with phrases like \textquotedblleft\raisebox{-0.2\height}{\includegraphics[height=1.3em]{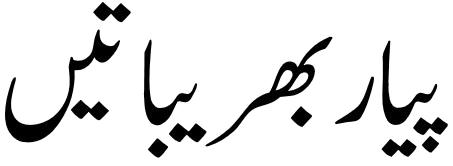}}\textquotedblright\ , meaning \textquotedblleft loving words\textquotedblright), where \textquotedblleft\raisebox{-0.2\height}{\includegraphics[height=1.1em]{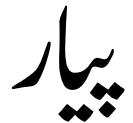}}\textquotedblright\ (pyaar, meaning \textquotedblleft love\textquotedblright), \textquotedblleft\raisebox{-0.2\height}{\includegraphics[height=1em]{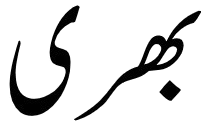}}\textquotedblright\ (bhari, meaning \textquotedblleft filled with\textquotedblright), and \textquotedblleft\raisebox{-0.2\height}{\includegraphics[height=1.2em]{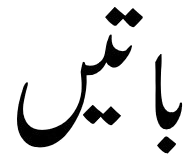}}\textquotedblright\ (batein, meaning \textquotedblleft words\textquotedblright) are merged without spaces. Without accurate tokenization, Urduhack may fail to separate these elements, potentially missing the emotional nuance in the phrase. To address these limitations, we conduct additional manual review during annotation, which enhances accuracy and maintains the dataset’s quality for MNER research. Urduhack thus establishes a solid foundation for Urdu tokenization, with supplementary corrections further ensuring its reliability across diverse text types.

\subsection{Data Annotation}
The data annotation process is a critical step in preparing the Twitter2015-Urdu dataset for MNER, focusing on accurately labeling named entities in Urdu-translated text to ensure consistency and precision. The initial annotation phase leverages existing labels in the Twitter2015 English dataset as a reference, guiding entity categories and boundaries. Annotators use these English annotations as a baseline, adhering closely to predefined guidelines to maintain alignment between the English and Urdu datasets, with adjustments made to accommodate Urdu-specific linguistic structures. For instance, certain entities in English, like \textquotedblleft White House\textquotedblright, which signifies the U.S. president’s residence, is translated as \textquotedblleft\raisebox{-0.2\height}{\includegraphics[height=1em]{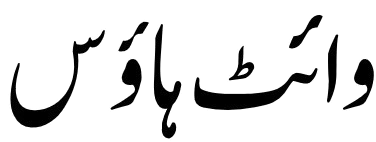}}\textquotedblright\ in Urdu and consistently annotated as a Location (LOC) to reflect its function rather than a literal translation. This approach maintains structural alignment between the datasets while adapting to the unique nature of the Urdu language.

To ensure annotation quality and consistency, two native Urdu-speaking annotators were recruited and instructed to label named entities in accordance with the predefined guidelines. A random subset of 500 tweets was selected for this purpose to mitigate potential annotation bias. To assess the reliability of the annotation process, inter-annotator agreement (IAA) was measured using Cohen’s Kappa coefficient. The resulting Kappa score of 0.773 indicates a substantial level of agreement between the annotators, thereby demonstrating the robustness and reliability of the annotation procedure. This further supports the dataset’s suitability for downstream Urdu MNER tasks. The confusion matrix between the two annotators is provided in Table~\ref{tab:kappa_matrix}, highlighting the points of agreement and discrepancies.

\begin{table}[ht]
\centering
\resizebox{\columnwidth}{!}{
\begin{tabular}{|l|c|c|c|c|c|c|}
\hline
\textbf{Annotator A $\downarrow$ / Annotator B $\rightarrow$} & \textbf{PER} & \textbf{LOC} & \textbf{ORG} & \textbf{OTHER} & \textbf{O} \\
\hline
\textbf{PER} & \textbf{360} & 60 & 30 & 20 & 15 \\
\hline
\textbf{LOC} & 40 & \textbf{330} & 45 & 35 & 35 \\
\hline
\textbf{ORG} & 25 & 35 & \textbf{70} & 10 & 8 \\
\hline
\textbf{OTHER} & 15 & 30 & 12 & \textbf{80} & 12 \\
\hline
\textbf{O (Non-entity)} & 15 & 30 & 18 & 15 & \textbf{7605} \\
\hline
\end{tabular}
}\caption{Reorganized Confusion Matrix of Kappa Statistic for Entity Annotation}
\label{tab:kappa_matrix}
\vspace{-0.3in}  
\end{table}

\subsection{Quality Control and Validation}
A self-review quality control process was implemented to ensure annotation consistency and reliability. Each annotator reviewed their own work by comparing their annotations to the predefined guidelines, which enabled them to detect and correct any initial errors while ensuring consistency across the dataset. The quality control approach establishes a solid foundation for data accuracy, making the dataset well-suited for MNER research. By closely aligning with the structure and categories of the English dataset, the annotation process effectively adapts to Urdu’s linguistic context.

\subsection{Dataset Characteristics}
The Twitter2015-Urdu dataset encompasses 8,257 text-image pairs, meticulously annotated to provide a robust resource for MNER research. The dataset features a comprehensive distribution of entity types across training, validation, and test sets. Specifically, it includes 2,255 persons (PER) in the training set, 558 in the validation set, and 1,939 in the test set; 2,076 locations (LOC) in the training set, 529 in the validation set, and 1,781 in the test set; 897 organizations (ORG) in the training set, 240 in the validation set, and 825 in the test set; and 946 miscellaneous (MISC) entities in the training set, 226 in the validation set, and 673 in the test set. This detailed and balanced distribution ensures the dataset's utility for comprehensive MNER research in Urdu. 
A comparative analysis between the original Twitter2015 dataset and the Twitter2015-Urdu dataset reveals several differences in entity proportions.

\begin{table}[htbp]
\centering
\renewcommand{\arraystretch}{1.4}  
\footnotesize  
\resizebox{\columnwidth}{!}{  
\begin{tabular}{|l|c|c|c|c|c|c|}
\hline
\textbf{Entity Type} & \multicolumn{3}{c|}{\textbf{Twitter-2015}} & \multicolumn{3}{c|}{\textbf{Twitter2015-Urdu}} \\ 
\cline{2-7}
& \textbf{Train} & \textbf{Val} & \textbf{Test} & \textbf{Train} & \textbf{Val} & \textbf{Test} \\
\hline
\textbf{PER}  & 2,217  & 552   & 1,816  & 2,255  & 558   & 1,939  \\
\textbf{LOC} & 2,091  & 522   & 1,697  & 2,076  & 529   & 1,781  \\
\textbf{ORG} & 928    & 247   & 839    & 897    & 240   & 825    \\
\textbf{MISC} & 940    & 225   & 726    & 946    & 226   & 673    \\
\hline
\textbf{Total Entities} & 6,176  & 1,546 & 5,078  & 6,174  & 1,553 & 5,218  \\
\hline
\textbf{Total Tweets} & 4,000  & 1,000 & 3,257  & 4,000  & 1,000 & 3,257  \\
\hline
\end{tabular}
}
\vspace{-1em}  
\caption{A detailed comparison of entity distribution between the original Twitter2015 dataset and the Twitter2015-Urdu dataset.}
\label{tab:comparison}
\vspace{-0.3in}  
\end{table}

Table~\ref{tab:comparison} provides a detailed comparison of entity distribution, highlighting the changes and adaptations made during translation.
These differences arise from translation ambiguities, annotation inconsistencies, and cultural and linguistic adaptations. Translation ambiguities arise when English entities have multiple possible equivalents in Urdu, such as \textquotedblleft UN\textquotedblright, which may appear as \textquotedblleft\raisebox{-0.2\height}{\includegraphics[height=1.2em]{images/urdutextimages/6.png}}\textquotedblright\ (United Nations) or transliterated as \textquotedblleft\raisebox{-0.2\height}{\includegraphics[height=0.9em]{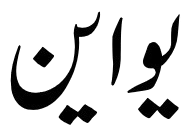}}\textquotedblright\ leading to inconsistencies. Annotation inconsistencies occur when entity boundaries differ across languages; for example, \textquotedblleft New York Times\textquotedblright \ is annotated as a single organization in English, but its Urdu equivalent, \textquotedblleft\raisebox{-0.2\height}{\includegraphics[height=1.1em]{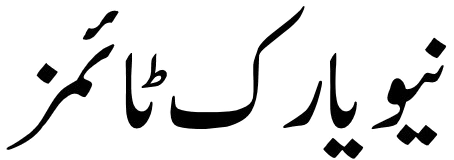}}\textquotedblright\ might be split into two tokens, requiring manual adjustments. Cultural adaptations are also necessary, as some entities carry different significance in Urdu contexts; for instance, certain English organizations like \textquotedblleft NBA\textquotedblright  \hspace{0.01cm} (National Basketball Association) are widely recognized in English-speaking countries and tagged as organization. However, in Urdu, where basketball may not be as popular, \textquotedblleft\raisebox{-0.2\height}{\includegraphics[height=1em]{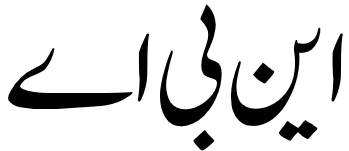}}\textquotedblright\ (NBA) might not hold the same recognition, and could sometimes be misinterpreted or omitted altogether.These factors highlight the complexities of translating and adapting datasets across languages, emphasizing the importance of careful review and quality control to maintain integrity and relevance.
\section{METHODOLOGY}
\subsection{PROBLEM FORMULATION}
MNER is a complex task that involves identifying and classifying named entities in text using additional contextual information provided by accompanying images. Given an input text-image pair $(T, V)$, the objective is to extract and classify a set of named entities into predefined categories such as PER, ORG, LOC, and MISC. Similar to existing work in MNER, this task is approached as a sequence labeling problem. For a given text $T=( w_1, w_2, \dots, w_n )$, where $w_i \in T$ denotes the $i^{\text{th}}$ word in the sentence, the goal is to predict a sequence of labels $Y=( y_1, y_2, \dots, y_n )$ following the standard BIO (Begin, Inside, Outside) tagging schema. The accompanying image $V$ provides visual context to enhance the disambiguation and classification of these entities.
\begin{figure*}[ht!]
    \centering
    \includegraphics[width=0.8\textwidth, height=0.9\textwidth]{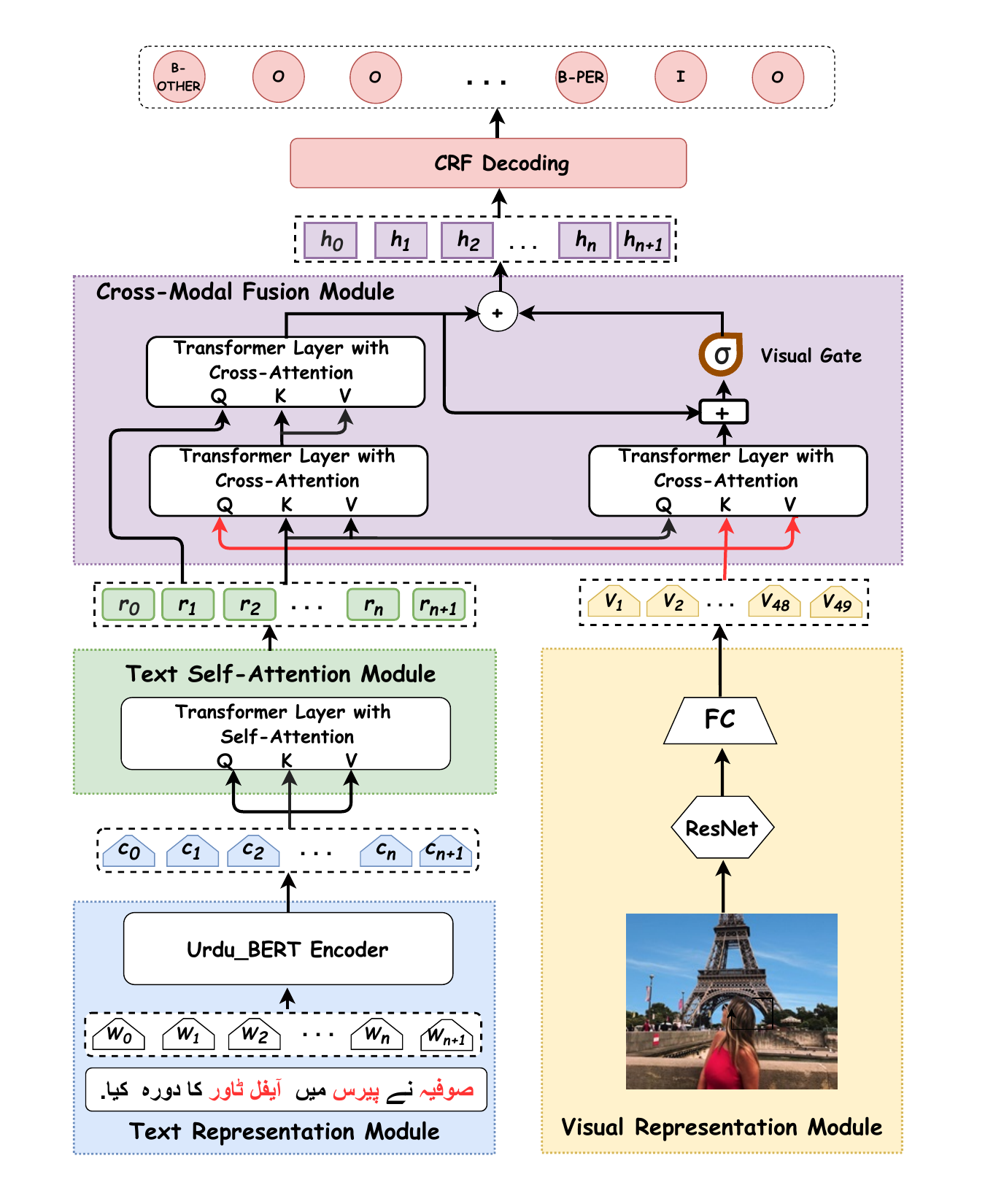}
        \caption{Overall Architecture of U-MNER Framework.}
    \label{fig2}
    \vspace{-0.1in}  
\end{figure*}
\subsection{OVERALL ARCHITECTURE}
The architecture of the proposed U-MNER framework is illustrated in Figure~\ref{fig2}, comprising five principal modules: (1) Text Representation Module, (2) Visual Representation Module, (3) Text Self-Attention Module, (4) Cross-Modal Fusion Module, and (5) CRF Decoder. This design is crafted to harness both textual and visual modalities, thereby enhancing NER performance. Each module contributes to a systematic processing of the input data, facilitating multimodal integration and accurate entity prediction.

The process begins with the Text Representation Module, where Urdu-BERT~\cite{malik2023contextual} is employed to generate contextualized word embeddings. These embeddings effectively capture the semantic intricacies and syntactic richness of the Urdu language. Concurrently, the Visual Representation Module utilizes ResNet to extract high-dimensional visual features from accompanying images, transforming them into a compact and meaningful representation using a fully connected (FC) layer. 

To capture long-range dependencies and contextual relationships within the text, the Text Self-Attention Module employs a Transformer Layer with Self-Attention. This mechanism allows the model to focus on relevant parts of the text, enabling it to capture both local and global linguistic dependencies. 

Subsequently, the Cross-Modal Fusion Module identifies and integrates visual regions pertinent to entities present in the text, enriching word representations via a multi-head cross-modal attention mechanism. A Visual Gate is introduced to selectively filters the visual features, allowing only the most relevant information to influence the text embeddings, which ensures that the model emphasizes meaningful multimodal interactions while disregarding irrelevant visual elements. These refined text and visual embeddings are then fused into a unified representation, bolstering the model’s capacity to disambiguate entities effectively. 

In the final stage, the CRF Decoding Module assigns entity labels to each word by modeling both local dependencies captured by the BiLSTM network and the global sequence structure through the CRF layer. This comprehensive approach yields precise and contextually informed named entity recognition by effectively synthesizing textual and visual data within a unified framework.

\subsection{TEXT REPRESENTATION MODULE}
To capture varying contextual meanings for the same word, we utilize the contextual embeddings provided by Urdu-BERT as our sentence encoder. In line with the preprocessing approach described by Devlin et al.~\cite{devlin2019bert}, each sentence is modified by adding two special tokens [CLS] at the beginning and [SEP] at the end. Formally, let $T'=(w_0,w_1,w_2,\dots,w_{n+1})$, represent the modified input sentence, where $w_0$ and $w_{n+1}$  represents the [CLS] and [SEP] tokens, respectively. The initial word embeddings are denoted as $E=(e_0,e_1,e_2,\dots,e_{n+1} )$, with each $e_0$ computed as the sum of the word, segment, and position embeddings for token $w_i$. These embeddings $E$ are then passed through the Urdu-BERT encoder, producing contextualized representations $C=(c_0,c_1,c_2,\dots,c_{n+1})$, where each $c_i \in \mathbb{R}^d$ is the contextualized representation for $e_i$.

\subsection{VISUAL REPRESENTATION MODULE}
We employ ResNet, a leading CNN model in image recognition, due to its effectiveness in capturing rich feature representations from images. Specifically, we use the output from the final convolutional layer of a pretrained 152-layer ResNet to represent each image. This layer divides each input image into 49 visual blocks, arranged in a $7 \times 7$ grid, with each block represented by a 2048-dimensional feature vector. For a given input image $V$, we first resize it to $224 \times 224$ pixels to maintain consistency, then extract visual features through ResNet, resulting in $U = \{u_1, u_2, u_3, \dots, u_{49}\}$, where each $u_i$  is the 2048-dimensional vector for the $i^{\text{th}}$ visual block. To align these visual features with the space of the word embeddings, we apply a fully connected layer, projecting $U$ into a lower-dimensional space. This transformation is expressed as $V = W_u^T U$, where $W_u \in \mathbb{R}^{2048 \times d}$ is the learned weight matrix. Consequently, $V = \{v_1, v_2, v_3, \dots, v_{49}\}$ represents the set of transformed visual embeddings, enabling seamless integration with text representations.

\subsection{TEXT SELF-ATTENTION MODULE}
To effectively capture intricate linguistic structures and contextual nuances within textual data, our framework utilizes a Transformer Layer with Self-Attention. This layer enables the model to process both local and global context by dynamically weighing the importance of each word relative to others in the sentence. Such capabilities are crucial for accurately identifying named entities, as the meaning of an entity often relies on the broader context.
The Transformer Layer receives contextualized representations from Urdu-BERT, represented as a sequence $C = (c_0, c_1, c_2, \dots, c_{n+1})$, where each $c_i$ captures the semantic context of a word. Using the self-attention mechanism, the model computes relationships between words, determining their relevance to one another. Specifically, the self-attention for each word representation $c_i$ is calculated as follows:

\begin{equation}
\text{Attention}(Q, K, V) = \text{softmax}\left(\frac{QK^T}{\sqrt{d_k}}\right)V
\label{eq:attention}
\end{equation}
Where $Q$,$K$ and $V$ represent the query, key, and value matrices derived from $C$, and $d_k$ is the dimension of the key vectors. This operation yields a sequence of enriched embeddings that encapsulate contextual relationships within the sentence.

These enriched embeddings are then processed by a feed-forward network, producing the final output $R = (r_0, r_1, r_2, \dots, r_{n+1})$, where each $r_i$ is further refined to capture both immediate and long-range dependencies. These final embeddings $R$ serve as a robust foundation for integrating visual data and enable more accurate, context-aware entity recognition. By leveraging the capabilities of the Transformer Layer with Self-Attention, our framework adeptly addresses the linguistic complexities of Urdu, thereby improving the accuracy of named entity classification.
\subsection{CROSS-MODAL FUSION MODULE}
Social media texts are often concise and ambiguous. Incorporating visual context provides additional cues that can help resolve ambiguities and enhance entity classification accuracy. To achieve this, we adopt the Cross-Modal Fusion Module based on the approach outlined in ~\cite{yu2020improving}, which integrates image-aware word representations and word-aware visual representations. This fusion enables more accurate recognition of named entities by leveraging both textual and visual modalities. 
\subsubsection{IMAGE-AWARE WORD REPRESENTATION}
In our framework, we enhance the word representations by integrating visual context through a Cross-Modal Transformer (CMT) layer. The primary goal is to refine the textual embeddings by incorporating information from the associated visual data. We achieve this by applying a multi-head cross-modal attention mechanism, where visual embeddings \( V \in \mathbb{R}^{d \times 49} \) are used as queries, and textual embeddings \( R \in \mathbb{R}^{d \times (n+1)} \) act as keys and values. This mechanism enables the model to learn relationships between visual features and text, allowing the representation of each word to be informed by the visual context.
The cross-modal attention is computed as follows:

\begin{equation}
\begin{split}
\text{MH-CA}(V, R) = W'\bigg[ \sum_{i=1}^{m} \text{softmax}\Bigg( 
\frac{[W_i^Q V]^T [W_i^K R]}{\sqrt{d/m}} \Bigg) \\
\cdot [W_i^V R]^T \bigg]^T
\end{split}
\end{equation}

\noindent The learned weight matrices \( W_i^Q \), \( W_i^K \), and \( W_i^V \in \mathbb{R}^{d/m \times d} \) correspond to the query, key, and value for each attention head, respectively, where \( m \) denotes the number of attention heads. The softmax function normalizes the attention scores, allowing the model to compute a weighted sum of the visual embeddings \( (W_i^V R)^T \). This process enables the model to capture the relationships between the textual input and its associated visual context, enhancing the word representations with relevant visual cues.

After the attention mechanism, the output is further refined using feed-forward network(\(\text{FFN}\)) and layer normalization(\(\text{LN}\)), as described in the following equation:
\begin{equation}\quad \tilde{P} = \text{LN}(V + \text{MH-CA}(V, R)),
\end{equation}
\begin{equation}
P = \text{LN}(\tilde{P} + \text{FFN}(\tilde{P})) \quad \end{equation}

\noindent Here, \( P = (p_0, p_1, p_2, \dots, p_{n+1}) \) represents the refined word embeddings enriched with both textual and visual information. \( \tilde{P} \), on the other hand, denotes the intermediate representation before the application of the feed-forward network (\(\text{FFN}\)). This process enables the integration of multimodal information, enhancing entity recognition tasks by incorporating relevant visual cues effectively. 

\textbf{Coupled CMT Layer:}
In the initial CMT layer, visual representations are used as queries, leading to a misalignment between the visual blocks and input words. To correct this, the Coupled CMT Layer reverses the roles: textual embeddings $R$ serve as queries, while  the output from the previous CMT layer $P$ act as keys and values. This ensures that the image-aware word representations align with the corresponding words in the input text. The Coupled CMT layer generates the final set of image-aware word representations, denoted as $A = (a_0, a_1, a_2, \dots, a_{n+1})$, where each $a_i$ captures the relevant visual context for the corresponding word $w_i$, enhancing the model’s ability to accurately recognize and classify named entities.

\subsubsection{WORD-AWARE VISUAL REPRESENTATION}
To align visual features with their corresponding textual elements, we use a CMT layer that generates word-aware visual representations. Here, textual embeddings $R$ are treated as queries, and visual embeddings $V$ act as keys and values. This alignment ensures that each word is associated with its most relevant visual blocks, resulting in a nuanced multimodal representation.
The cross-modal attention mechanism is similar to that used for image-aware word representations, producing the word-aware visual representations $Q = (q_0, q_1, q_2, \dots, q_{n+1})$, where each $q_i$ corresponds to the visual context most relevant to the word $w_i$.

\subsubsection{VISUAL GATE} 
Incorporating visual data requires a selective approach to ensure that only relevant visual information contributes to word representations. Previous studies ~\cite{zhang2018coattention,lu2018visual} have observed that aligning function words—such as the Urdu equivalents of \textquotedblleft\raisebox{-0.2\height}{\includegraphics[height=0.8em]{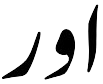}}\textquotedblright\ (and) or \textquotedblleft\raisebox{-0.2\height}{\includegraphics[height=1em]{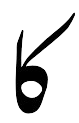}}\textquotedblright\ —with visual features is often unnecessary, as these words generally lack visually interpretable meanings. To address this, our framework introduces a visual gate mechanism that dynamically regulates the influence of visual features on each word's representation, allowing the model to prioritize meaningful multimodal interactions.
The visual gate is formulated by combining information from both the word representations $A$ and the visual representations $Q$, as follows:
\begin{equation}
g = \sigma\left(W_a^T A + W_q^T Q\right)
\label{eq:g}
\end{equation}
where $W_a, W_q \in \mathbb{R}^{d \times d}$ are learned weight matrices, and $\sigma$ represents the element-wise sigmoid activation function. This gating mechanism modulates the integration of visual context, enabling the model to selectively incorporate visual features where they add meaningful context.
Using the gate output, the final word-aware visual representation $B$ is calculated as:    
\begin{equation}
B = g \odot Q
\label{eq:B}
\end{equation}
where $\odot$ denotes element-wise multiplication. This approach ensures that only visually relevant features are incorporated to enhance each word's representation, optimizing the model’s ability to capture meaningful multimodal interactions. The selective integration of visual information allows the model to focus on contextually appropriate features, thereby enhancing accuracy in MNER tasks, especially in cases where visual and textual cues need to be aligned effectively.

\subsection{CRF DECODER}
After obtaining the image-aware word representations $A$ and the word-aware visual representations $B$, we concatenate these vectors to form the final hidden representations $H = \{h_0, h_1, h_2, \dots, h_{n+1}\}$, 
\begin{equation}
h_i = \text{Concat}(a_i, b_i) \quad \text{for} \quad i = 0, 1, 2, \dots, n+1
\label{eq:concat}
\end{equation}
here, $h_i \in \mathbb{R}^{2d}$ captures both textual and visual contexts, ensuring that the multimodal information is integrated into a single representation for each word.

\subsubsection{BILSTM PROCESSING}
The concatenated representations $H$ are then processed by a Bidirectional Long Short-Term Memory (BiLSTM) network, which produces hidden states $(\overrightarrow{h_i}, \overleftarrow{h_i})$ for each word:
\begin{equation}
\overrightarrow{h}_i = \text{LSTM}_{\text{forward}}\left(h_i, \overrightarrow{h}_{i-1}\right)
\label{eq:lstm-forward}
\end{equation}

\begin{equation}
\overleftarrow{h}_i = \text{LSTM}_{\text{backward}}\left(h_i, \overleftarrow{h}_{i-1}\right)
\label{eq:lstm-backward}
\end{equation}
The combined BiLSTM output for each word is then:
\begin{equation}
h_i^{\text{BiLSTM}} = \text{Concat}\left(\overrightarrow{h}_i, \overleftarrow{h}_i\right)
\label{eq:bilstm}
\end{equation}
This step captures both forward and backward dependencies, enriching the representation of each word with context from the entire sequence.

\subsubsection{CRF LAYER}
The hidden states $h_i^{\text{BiLSTM}}$ are passed to a Conditional Random Field (CRF) layer to assign entity labels $Y = \{y_0, y_1, y_2, \dots, y_{n+1}\}$ to each word, considering the entire sequence. The CRF layer computes the probability of a label sequence $Y$ given the input sequence $(T,V)$ as:
\begin{equation}
P\left(Y \mid (T, V)\right) = \frac{\exp\left(\text{score}(H, Y)\right)}{\sum_{Y' \in \mathcal{Y}} \exp\left(\text{score}(H, Y')\right)}
\label{eq:probability}
\end{equation}

\textbf{Score Function:} The score function $\text{score}(H, Y)$ calculates the combined likelihood of a sequence based on emission and transition scores:
\begin{equation}
\text{score}(H, Y) = \sum_{i=0}^n \mathcal{T}_{y_i, y_{i+1}} + \sum_{i=0}^n E\left(h_i^{\text{BiLSTM}}, y_i\right)
\label{eq:score}
\end{equation}
here, $\mathcal{T}(y_i, y_{i+1})$ denotes the transition score between consecutive labels $y_i$ and $y_{i+1}$, and the emission score $E(h_i^{\text{BiLSTM}}, y_i)$ is calculated as:
\begin{equation}
E\left(h_i^{\text{BiLSTM}}, y_i\right) = W_{y_i}^{\text{MNER}} \cdot h_i^{\text{BiLSTM}}
\label{eq:emission}
\end{equation}
where $W_{y_i}^{\text{MNER}} \in \mathbb{R}^{2d}$ is the weight vector specific to the label $y_i$. This approach allows the CRF to leverage both the local dependencies captured by the BiLSTM (emission scores) and the global sequence structure (transition scores), leading to more precise and contextually aware named entity recognition.

\textbf{Training Loss:} The training loss for the CRF layer is based on negative log-likelihood, which seeks to maximize the probability of correct label sequences:
\begin{equation}
L = -\log\left(P\left(Y \mid (S, V)\right)\right)
\label{eq:loss}
\end{equation}
This loss function ensures that the CRF learns to assign high scores to correct label sequences during training, thus enhancing the accuracy and coherence of the model's predictions.
\section{EXPERIMENTS}
This section presents a comprehensive evaluation of various NER models on the Twitter15-Urdu dataset, comparing text-based and multimodal approaches to highlight the effectiveness of incorporating visual context.
\subsection{EXPERIMENT SETTINGS}
\subsubsection{DATASETS}
For our experiments, we utilized the Twitter2015-Urdu dataset, a multimodal dataset for Urdu MNER. This dataset contains tweets annotated with four entity types: Person (PER), Location (LOC), Organization (ORG), and Miscellaneous (MISC). Its construction accounts for the unique linguistic and visual features of Urdu social media posts.
\subsubsection{BASELINES}
To evaluate the performance of our U-MNER model, we compared it against several benchmark models in both text-only and text+vision categories:\\\\
\noindent\textbf{Text-Based Models:}
\begin{itemize}
    \item \textbf{LSTM-CRF}~\cite{zeng2017lstmcrf}: A traditional sequence labeling model that uses Long Short-Term Memory (LSTM) networks to capture sequential dependencies in text, followed by a Conditional Random Field (CRF) layer to model the dependencies between output labels and ensure a valid sequence of predictions.
    
    \item \textbf{BiLSTM-CRF}~\cite{huang2015bilstmcrf}: Extends the LSTM-CRF model by incorporating bidirectional LSTM layers, allowing the model to capture context from both past and future directions, thereby improving the quality of the learned representations for each word in the sequence.
    
    \item \textbf{HBiLSTM-CRF}~\cite{lample2016neural}: Enhances the BiLSTM-CRF architecture by adding hierarchical bidirectional LSTMs that include character-level representations. This allows the model to capture richer morphological and syntactical features, which are particularly useful for recognizing named entities.
    
    \item \textbf{CNN-BiLSTM-CRF}~\cite{ma2016end}: Combines Convolutional Neural Networks (CNNs) for extracting character-level features with BiLSTM and CRF layers. The CNNs capture local patterns in characters, while BiLSTM and CRF layers handle sequence labeling, providing a robust model for NER tasks.
    
    \item \textbf{BERT}~\cite{devlin2019bert}: Utilizes the Bidirectional Encoder Representations from Transformers (BERT) model, a pre-trained Transformer architecture that generates contextualized word embeddings. For NER, a softmax layer is added on top to predict entity labels based on these embeddings.
    
    \item \textbf{BERT-CRF}~\cite{yu2020improving}: Enhances the BERT model by incorporating a CRF layer on top of the Transformer layers. This allows the model to leverage BERT's powerful contextual embeddings while capturing label dependencies, leading to more accurate sequence labeling.
\end{itemize}
\begin{table*}[htbp]
\centering
\small 
\renewcommand{\arraystretch}{1.2} 
\resizebox{\textwidth}{!}{ 
\begin{tabular}{|c|c|S[table-format=2.2]|S[table-format=2.2]|S[table-format=2.2]|S[table-format=2.2]|S[table-format=2.2]|S[table-format=2.2]|S[table-format=2.2]|}
\toprule
\textbf{Modality} & \textbf{Methods} & \multicolumn{4}{c|}{\textbf{Single Type (F1)}} & \multicolumn{3}{c|}{\textbf{Overall}} \\
 &  & \textbf{PER} & \textbf{LOC} & \textbf{ORG} & \textbf{MISC} & \textbf{P} & \textbf{R} & \textbf{F1} \\
\midrule
\multirow{6}{*}{\textbf{Text}} 
 & LSTM-CRF & 63.36 & 56.90 & 34.85 & 16.42 & 56.44 & 45.46 & 50.36 \\
 & BiLSTM-CRF & 64.97 & 57.82 & 34.28 & 19.99 & 58.08 & 46.95 & 51.93 \\
 & HBiLSTM-CRF & 67.83 & 58.71 & 30.92 & 18.05 & 57.10 & 49.04 & 52.76 \\
 & CNN-BiLSTM-CRF & 58.23 & 48.51 & 28.61 & 19.47 & 56.96 & 45.23 & 50.42 \\
 & BERT & 72.01 & 67.74 & 39.89 & 15.36 & 62.08 & 57.22 & 59.55 \\
 & BERT-CRF & 72.85 & 68.16 & 41.29 & 16.65 & 63.32 & 56.87 & 59.92 \\
\midrule
\multirow{6}{*}{\textbf{Text+Vision}} 
 & UMT & 68.03 & 64.77 & 40.71 & 20.11 & 52.80 & 55.41 & 54.07 \\
 & RpBERT & 72.47 & 68.48 & 39.07 & 21.11 & 63.22 & 57.69 & 60.82 \\
 & HVPNET & 72.28 & 70.47 & 43.47 & \textbf{25.80} & 62.93 & 60.03 & \textbf{60.92} \\
 & MAF & 73.75 & 69.32 & 47.09 & 25.30 & 59.85 & 61.30 & 60.57 \\
 & MGCMT & 69.71 & 65.31 & 42.60 & 24.36 & 57.28 & 55.57 & 56.41 \\
 & \textbf{U-MNER (Ours)} & \textbf{73.83} & \textbf{70.71} & \textbf{47.91} & 23.32 & \textbf{63.27} & \textbf{62.24} & \textbf{62.75} \\
\bottomrule
\end{tabular}
}
\caption{Performance Comparison of Text-Based and Multimodal NER Models on the Twitter15-Urdu Dataset.}
\label{tab:performance}
\vspace{-0.2in}  
\end{table*}

\noindent\textbf{Text+Vision Models:}
\begin{itemize}
    \item \textbf{UMT}~\cite{yu2020improving}: Unified Multimodal Transformer that jointly models text and vision inputs. UMT leverages Transformer architectures to seamlessly integrate multimodal information, providing a unified approach to understanding and recognizing entities in social media posts.
    
    \item \textbf{RpBERT}~\cite{sun2021rpbert}: Extends the RoBERTa model with multimodal features, enabling it to handle both text and vision inputs. RpBERT enriches entity recognition by integrating visual cues with robust textual representations, resulting in better performance on multimodal NER tasks.
    
    \item \textbf{HVPNET}~\cite{chen2022good}:
    Hierarchical Visual Processing Network designed for aligning and integrating text and visual features. HVPNET processes visual data hierarchically, aligning it with textual information to enhance entity recognition accuracy through better multimodal context understanding.
    
    \item \textbf{MAF}~\cite{xu2022maf}: Uses Multimodal Attention Fusion to combine text and image data. The model applies attention mechanisms to selectively integrate information from both modalities, improving the accuracy of named entity recognition by leveraging complementary visual context.

    \item \textbf{MGCMT}~\cite{LIU2024103546}: Utilizes Multi-Granularity Cross-Modal Transformers to integrate textual and visual data at multiple levels of granularity. This approach captures fine-grained interactions between modalities, enhancing overall multimodal understanding.
    
    \end{itemize}
\subsubsection{EVALUATION MATRIX}
To assess the performance of the MNER models, we utilize the F1 score for each entity type, along with the overall precision (P), recall (R), and F1 score (F1). These metrics provide a comprehensive evaluation of model accuracy and effectiveness. 
\subsubsection{IMPLEMENTATION DETAILS}
Our experiments were conducted using PyTorch on a system equipped with an NVIDIA GeForce RTX 4090 GPU, running on NVIDIA-SMI 535.161.07 with CUDA version 12.2. This powerful GPU setup, with 24,564 MiB of available memory, enables efficient processing of complex models and large datasets, significantly enhancing computational performance for training and inference. We carefully tuned the hyperparameters to ensure optimal performance across both unimodal and multimodal approaches, particularly focusing on our U-MNER model for the Twitter15-Urdu dataset. For text representations, we used the cased BERT base model, which was fine-tuned during training to capture the linguistic nuances of Urdu. This fine-tuning allowed the model to adapt to the specific linguistic characteristics prevalent in Urdu social media texts. For visual representations, we employed a pre-trained ResNet-152 model with fixed weights, preserving the integrity of the visual features and ensuring effective multimodal integration. Key hyperparameters were set as follows: the learning rate was 5e-5, the dropout rate was 0.1, and the batch size was 16. The maximum sentence length was set to 128 tokens, and we utilized 12 cross-modal attention heads to enhance interactions between textual and visual modalities. The tradeoff parameter $\lambda$ was set to 0.5. These settings were determined through a grid search on the development set, exploring various combinations of learning rates [1e-5, 1e-4], dropout rates [0.1, 0.5], and tradeoff parameters [0.1, 0.9]. This meticulous tuning ensured a fair comparison across models and highlighted the effectiveness of our approach in integrating multimodal data for robust and context-aware named entity recognition in Urdu social media contexts.
\subsection{EXPERIMENTAL RESULTS AND ANALYSIS}
To effectively verify the advantages of our proposed U-MNER model, we compared it against benchmark models using the same experimental dataset distribution. The results of these experiments are presented in Table~\ref{tab:performance}.

\subsubsection{\textbf{Analysis of Text-based NER Methods}}
From the Table \ref{tab:performance}, it is clear that BERT-based methods significantly outperform LSTM-based methods in terms of F1 score. Specifically, on the Twitter2015-Urdu dataset, the BERT model achieved an F1 score of 59.55\%, showing an absolute improvement of 9.13\% over the 50.42\% achieved by CNN-BiLSTM-CRF. This indicates that the pre-trained BERT model is better suited for handling scenarios with insufficient text information and noisy data compared to CNN and LSTM models. Additionally, the BERT-CRF model further improves the F1 score to 59.92\%, showing a slight enhancement over BERT alone due to the CRF layer's ability to model label dependencies effectively.
\subsubsection{\textbf{Analysis of Multimodal NER Methods}}
 Models incorporating visual modality information consistently outperform their text-only counterparts. HVPNET outperformed BiLSTM-CRF by 9.59\%, while RpBERT achieved 0.90\% higher than BERT-CRF. These results highlight that visual information can address the ambiguity and fuzziness in textual data, enriching the context of social media posts and significantly contributing to improved named entity recognition.
\subsubsection{\textbf{Comparison with All Other Multimodal NER Methods}}
Our U-MNER model achieved a state-of-the-art F1 score of 62.75\% on the Twitter2015-Urdu dataset, outperforming all other evaluated models. Specifically, U-MNER showed an absolute improvement of 1.83\% over HVPNET, the previous best-performing multimodal model. These results are a direct reflection of U-MNER’s ability to effectively leverage multimodal data for named entity recognition. This improvement underscores the effectiveness of our cross-modal fusion mechanism, visual gating strategy, and the tailored fine-tuning of the BERT model for the Urdu language, which together resolve ambiguities and enhance entity recognition.

The comparative results, presented in Table~\ref{tab:performance}, demonstrate that U-MNER not only outperforms text-only models but also offers consistent performance gains across most entity types, reinforcing its robustness. These results position U-MNER as the state-of-the-art model for Urdu MNER on this benchmark, achieving superior performance over both traditional text-only methods and prior multimodal models.
\subsection{Efficiency}
The computational efficiency of U-MNER, as shown in Table IV, reflects a trade-off between its complexity and performance. With 476.66 million parameters, U-MNER requires 1987.8 seconds for training and 39.05 seconds for inference. The high computational cost is due to its sophisticated architecture, which combines Urdu-BERT for text processing, ResNet for visual feature extraction, and a Cross-Modal Fusion Module that integrates both modalities using attention mechanisms. In contrast, MAF and HypNet, with fewer parameters (205.63 million and 167.37 million, respectively), are faster, requiring 774.99 and 511.48 seconds for training, and 19.43 and 2.28 seconds for inference. While these models are more computationally efficient, U-MNER’s higher computational cost is justified by its superior performance in multimodal named entity recognition tasks, making it a better choice for applications requiring higher accuracy and robustness despite the longer processing times.
\begin{table}[ht]
\centering
\resizebox{\columnwidth}{!}{
\begin{tabular}{|l|c|c|c|}
\hline
\textbf{Model $\downarrow$} & \textbf{Parameters (M)} & \textbf{Training Time} & \textbf{Inference Time} \\
\hline
\textbf{U-MNER} & 476.66 & 1987.8 & 39.05 \\
\hline
\textbf{MAF}    & 205.63 & 774.99 & 19.43 \\
\hline
\textbf{HvpNET} & 167.37 & 511.48 & 2.28 \\
\hline
\end{tabular}
}
\caption{Comparison of Computational Efficiency of U-MNER, MAF, and HvpNET}
\label{tab:efficiency_comparison}
\vspace{-0.3in}  
\end{table}

\begin{table}[htbp]  
\centering
\renewcommand{\arraystretch}{1.4}  
\footnotesize  

\resizebox{\columnwidth}{!}{  
\begin{tabular}{|l|c|c|c|}
\hline
\textbf{Methods}               & \textbf{Precision} & \textbf{Recall} & \textbf{F1 Score} \\ \hline
\textbf{U-MNER}              & \textbf{66.15}     & \textbf{59.68}  & \textbf{62.75}    \\ \hline
\textbf{w/o SA}                & 64.50             & 58.20           & 61.20            \\ 
\textbf{w/o MR}              & 63.80             & 57.70           & 60.60            \\ 
\textbf{w/o VG}                & 65.90             & 59.00           & 62.20            \\ 
\textbf{w/o SA \& MR}        & 62.30             & 55.90           & 58.90            \\ 
\textbf{w/o SA, MR \& VG}    & 61.50             & 54.80           & 58.00            \\ \hline
\end{tabular}
}
\caption{Ablation study of U-MNER showing the impact of removing Self-Attention (SA), Multimodal Representation (MR), and Visual Gate (VG) on Precision, Recall, and F1 Score.}
\label{tab:ablation}
\vspace{-0.3in}  
\end{table}
\subsection{ABLATION STUDY}
Table~\ref{tab:ablation} shows the results of ablation study to demonstrate the effectiveness of the two key components: the Text Self-Attention Module (SA) and the Cross-Modal Fusion Module (CMFM). To better demonstrate the functionality of the Cross-Modal Fusion Module, it is divided into two parts: (1) Multimodal Representation (MR) and (2) Visual Gate (VG). MR refers to the integration of image-aware word representations and word-aware visual representations, which enables the model to align and enhance textual and visual features for robust entity recognition. By systematically removing each component (or combinations of them), we can analyze the model's performance degradation. The removal of each component (or all) results in a varying degree of performance decline, indicating that all three components play indispensable roles in enhancing the overall model performance.

{\textbf{Without Self-Attention (w/o SA):}
The F1 score drops from 62.75\% to 61.20\%, indicating that SA improves the model’s understanding of long-range dependencies within the text. Without this, the model’s ability to contextualize entities weakens, leading to a slight performance decline.

{\textbf{Without Multimodal Representation (w/o MR):}
Removing MR results in an F1 score of 60.60\%, showing a clear decrease in performance. The absence of cross-modal attention impacts how well the model integrates visual data with textual information, underscoring its importance for multimodal tasks.

{\textbf{Without Visual Gate (w/o VG):}
The F1 score is reduced to 62.20\% when the Visual Gate is removed, indicating that the gate helps filter irrelevant visual information. However, its absence does not hurt the performance as significantly as removing SA or MR.

{\textbf{Without Both SA and MR (w/o SA \& MR):}
With the removal of both Self-Attention and Multimodal Representation, the F1 score drops to 58.90\%. This significant reduction shows the combined importance of handling long-range dependencies in text and using cross-modal interactions effectively.

{\textbf{Without All Components (w/o SA, MR \& VG):}
When all three components are removed, the F1 score declines further to 58.00\%. This shows the cumulative impact of losing the ability to process long-range text dependencies, cross-modal attention, and the filtering of irrelevant visual information, indicating the necessity of each component for optimal performance.

\begin{figure*}[ht!]
    \centering
    \includegraphics[width=7.16in]{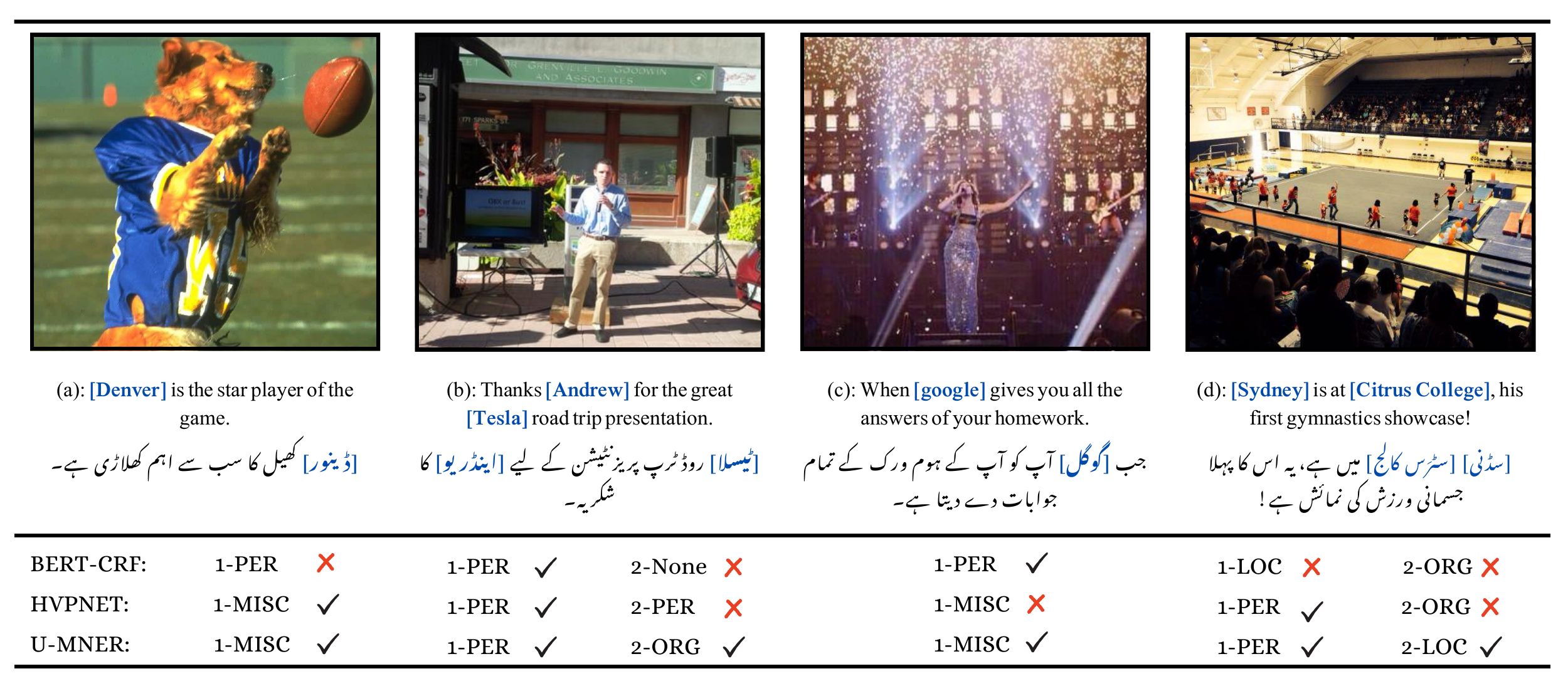} 
    \caption{Case study illustrating how visual data improves entity recognition by resolving ambiguities and enhancing classification in multimodal NER.}
    \label{Figure3}
\end{figure*}

\subsection{CASE STUDY}
To further demonstrate the effectiveness of our MNER framework, we present four typical cases (as shown in Figure~\ref{Figure3}) that highlight how incorporating visual information improves entity recognition and resolves ambiguities that a text-only model might struggle with. We also compare the predictions made by BERT-CRF, HVPNET, and our U-MNER model to showcase the differences in performance across models.

\textbf{Case (a): Resolving Ambiguity with Visual Cues:}
This case highlights how visual representations can help resolve ambiguity in entity classification. The tweet, \textquotedblleft\raisebox{-0.2\height}{\includegraphics[height=1.5em]{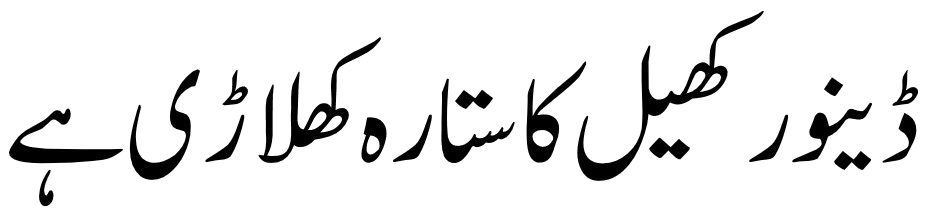}}\textquotedblright\ (Denver is the star player of the game), presents ambiguity, as the text alone doesn't clarify whether \textquotedblleft\raisebox{-0.2\height}{\includegraphics[height=1em]{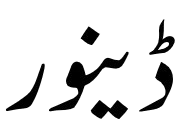}}\textquotedblright\ (Denver) is a person or something else. The text-only model BERT-CRF misclassifies \textquotedblleft\raisebox{-0.2\height}{\includegraphics[height=1em]{images/urdutextimages/23.png}}\textquotedblright\ (Denver) as a person based on common associations. However, both HVPNET~\cite{chen2022good} and our U-MNER model, which incorporate visual data, correctly classify \textquotedblleft\raisebox{-0.2\height}{\includegraphics[height=1em]{images/urdutextimages/23.png}}\textquotedblright\ (Denver) as a miscellaneous entity, recognizing from the image that Denver refers to a dog catching a ball. This case demonstrates the significance of visual information in resolving textual ambiguities.

\textbf{Case (b): Clarifying Ambiguous Entities:}
In this case, the ability of U-MNER to identify multimodal correlations is apparent. The entity \textquotedblleft\raisebox{-0.2\height}{\includegraphics[height=1em]{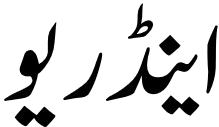}}\textquotedblright\ (Andrew) is correctly identified as a person by all models. However, BERT-CRF incorrectly identifies \textquotedblleft\raisebox{-0.2\height}{\includegraphics[height=1.1em]{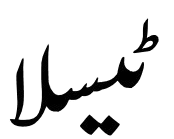}}\textquotedblright\ (Tesla) as a non-entity, lacking the visual context needed for proper classification. HVPNET also misclassifies \textquotedblleft\raisebox{-0.2\height}{\includegraphics[height=1.1em]{images/urdutextimages/25.png}}\textquotedblright\ (Tesla) as a person. In contrast, U-MNER effectively uses the visual cue of a Tesla car in the background to correctly classify \textquotedblleft\raisebox{-0.2\height}{\includegraphics[height=1.1em]{images/urdutextimages/25.png}}\textquotedblright\ (Tesla) as an organization. This case underscores the importance of aligning text and image data for accurate entity recognition, especially for ambiguous names.

\textbf{Case (c): Filtering Visual Noise:}
This case illustrates the ability of U-MNER to handle visual noise effectively. The tweet, \textquotedblleft\raisebox{-0.2\height}{\includegraphics[height=1.5em]{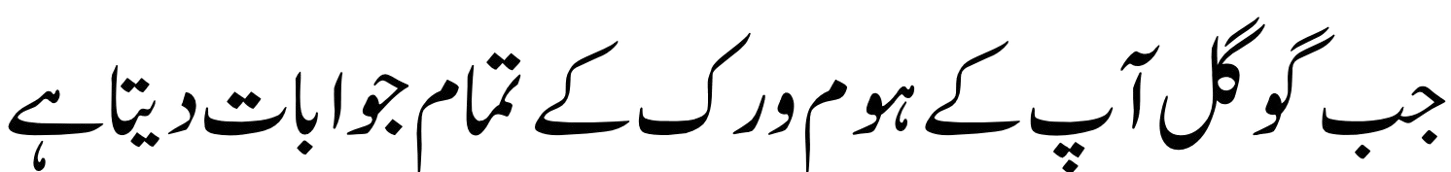}}\textquotedblright\ (When Google gives you all the answers to your homework), refers to (Google) as an organization. However, the accompanying image shows a concert, introducing irrelevant visual information. Despite the noise, BERT-CRF correctly identifies \textquotedblleft Google\textquotedblright \hspace{0.01cm} as an organization based solely on the text. HVPNET, overwhelmed by the visual context, fails to classify \textquotedblleft Google\textquotedblright \hspace{0.01cm} correctly. U-MNER, equipped with a Visual Gate mechanism, filters out the noise from the concert scene and correctly classifies \textquotedblleft Google\textquotedblright \hspace{0.01cm} as an organization, demonstrating its ability to balance multimodal information and handle noisy visual data.
\begin{figure*}[ht!]
     \centering
    \includegraphics[width=7.16in]{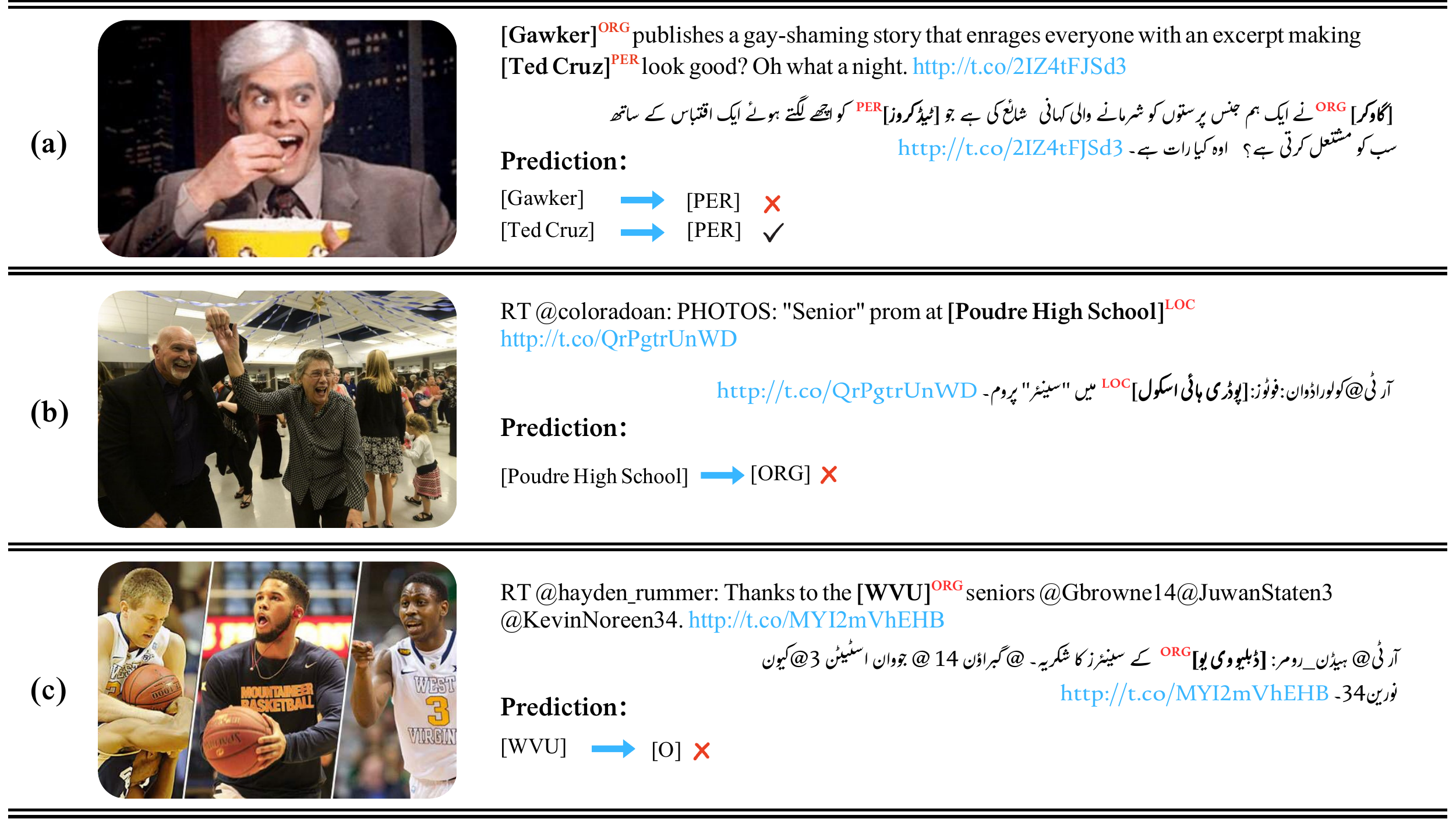} 
    \caption{Error analysis examples from U-MNER, showing misclassifications due to image-text mismatch, contextual ambiguity, and challenges with abbreviations.}
    \label{Figure5}
\end{figure*}

\textbf{Case (d): Addressing Location and Person Overlap:}
In this case, both \textquotedblleft\raisebox{-0.2\height}{\includegraphics[height=1.1em]{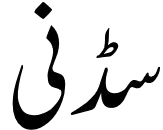}}\textquotedblright\ (Sydney) and \textquotedblleft\raisebox{-0.2\height}{\includegraphics[height=1.3em]{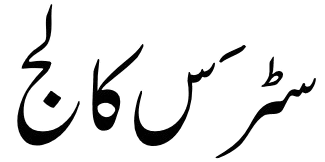}}\textquotedblright\ (Citrus College) are mentioned, leading to potential confusion between a person and a location. BERT-CRF misclassifies \textquotedblleft\raisebox{-0.2\height}{\includegraphics[height=1.1em]{images/urdutextimages/27.png}}\textquotedblright\ (Sydney) as a location and \textquotedblleft\raisebox{-0.2\height}{\includegraphics[height=1.3em]{images/urdutextimages/28.png}}\textquotedblright\ (Citrus College) as an organization due to the lack of contextual understanding. HVPNET, with its better multimodal integration, correctly identifies \textquotedblleft\raisebox{-0.2\height}{\includegraphics[height=1.1em]{images/urdutextimages/27.png}}\textquotedblright\ (Sydney) as a person and \textquotedblleft\raisebox{-0.2\height}{\includegraphics[height=1.3em]{images/urdutextimages/28.png}}\textquotedblright\ (Citrus College) as a location by effectively using both text and visual context. However, U-MNER, despite identifying \textquotedblleft\raisebox{-0.2\height}{\includegraphics[height=1.1em]{images/urdutextimages/27.png}}\textquotedblright\ (Sydney) as a person, misclassifies \textquotedblleft\raisebox{-0.2\height}{\includegraphics[height=1.3em]{images/urdutextimages/28.png}}\textquotedblright\ (Citrus College) as an organization. This misclassification likely stems from the model’s over-reliance on textual patterns, where terms like \textquotedblleft\raisebox{-0.2\height}{\includegraphics[height=1.3em]{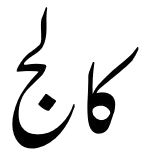}}\textquotedblright\ (College) are often associated with organizations. The model may have overlooked the broader visual clues indicating that \textquotedblleft\raisebox{-0.2\height}{\includegraphics[height=1.3em]{images/urdutextimages/28.png}}\textquotedblright\ (Citrus College) refers to a location, particularly within the context of a sports event at a campus. This case illustrates U-MNER's difficulty in dealing with entities that could fit multiple categories based on context.

Overall, the case study demonstrates U-MNER's ability to leverage multimodal data, improving accuracy by resolving ambiguities, enhancing classifications, and filtering visual noise. While outperforming text-only models, challenges remain in handling overlapping entity categories, highlighting the The CRF layer
enhances the model’s accuracy by considering the contextual dependencies between labels need for better multimodal alignment in complex scenarios.

\subsection{Error Analysis}
To better understand the limitations of the U-MNER framework, we performed an error analysis focusing on misclassified instances. As shown in Figure~\ref{Figure5}, one key challenge is the mismatch between the text and the visual modality, often leading to incorrect entity classification. For example (a), the term Gawker, which should be classified as an organization (ORG), is misclassified as a person (PER) due to the presence of a human figure in the image. This highlights the model's over-reliance on visual cues. Another issue arises from contextual ambiguity, particularly with short, noisy texts often seen on social media. In example (b), the entity Poudre High School is incorrectly labeled as an organization (ORG) instead of the correct location (LOC) label, despite clear textual context. In Urdu MNER, this misclassification stems from the model’s difficulty distinguishing between locations and institutions. The term High School in Urdu can refer to both a physical location and an institution, and the lack of additional context in the tweet makes it challenging for the model to identify it as a location. This issue is more prominent in short, context-poor texts like tweets, where the model may struggle with ambiguous terms. Example (c) demonstrates the model's difficulty with abbreviations and acronyms, such as WVU (West Virginia University), which is misclassified as a non-entity (O) instead of an organization (ORG). These errors likely stem from the inconsistent transliteration of abbreviations in Urdu and the lack of sufficient contextual information in brief social media posts.

\section{Conclusion and future work}
In this study, we present the Twitter2015-Urdu dataset, the first MNER dataset for Urdu. To establish benchmark baselines, we evaluate the dataset using both text-based and multimodal models. Furthermore, we propose the U-MNER framework, which leverages cross-attention mechanisms to integrate text and image information. Experimental results demonstrate that the U-MNER model achieves state-of-the-art performance on the Twitter2015-Urdu dataset, effectively capturing complex dependencies between modalities. 

This work provides a foundation for future research on Urdu MNER and offers valuable insights for exploring multimodal approaches in low-resource languages. U-MNER’s ability to process both text and visual content enables impactful applications in domains such as social media analysis, sentiment monitoring, and political discourse analysis. Its potential to bridge language gaps makes it particularly useful for advancing AI technologies in underserved regions. From an ethical standpoint, this research highlights considerations such as mitigating dataset bias and ensuring fair representation, especially for underrepresented groups. As the dataset relies on publicly available content, it ensures no private data is used.

For future work, we plan to expand the Twitter2015-Urdu dataset by including more diverse, real-world data from various domains, which will help improve the model's generalization capabilities. Additionally, we aim to explore the use of generative models to create synthetic data, further enhancing training and robustness. Finally, we will focus on reducing the computational complexity of the model to make it more efficient and scalable for deployment in real-time applications.

\bibliographystyle{IEEEtran} 
\bibliography{custom} 

\vspace{-2.5in}  
\begin{IEEEbiography}[{\includegraphics[width=1.1in,height=3in,clip,keepaspectratio]{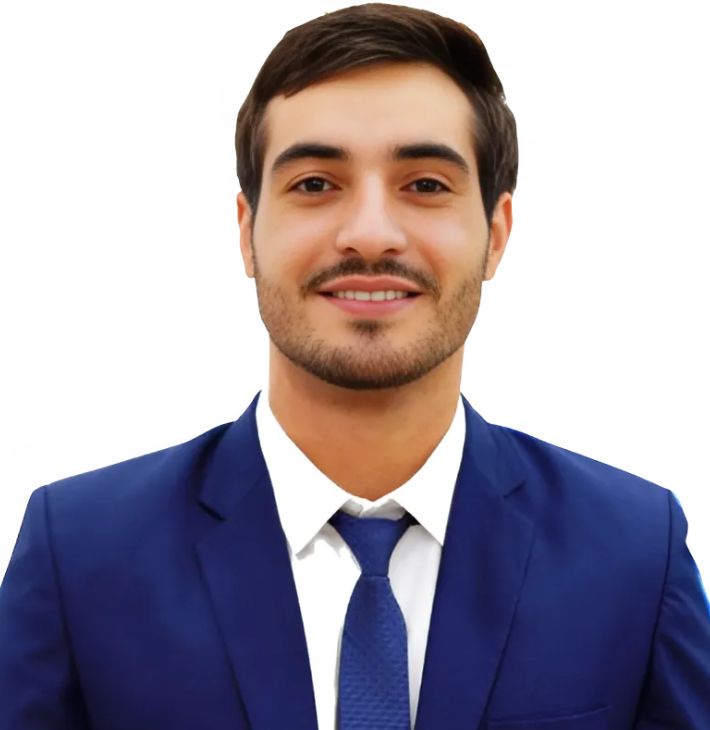}}]{\\Hussain Ahmad} is currently pursuing a Master’s degree in Information Science and Technology at Beijing University of Chemical Technology, China. He received his Master of Computer Science degree from the University of Chitral in 2022, graduating with distinction and ranking first in his class. His research interests include multimodal systems and natural language processing. 
\end{IEEEbiography}
\vspace{-2.5in}  
\begin{IEEEbiography}[{\includegraphics[width=1.1in,height=1.25in,clip,keepaspectratio]{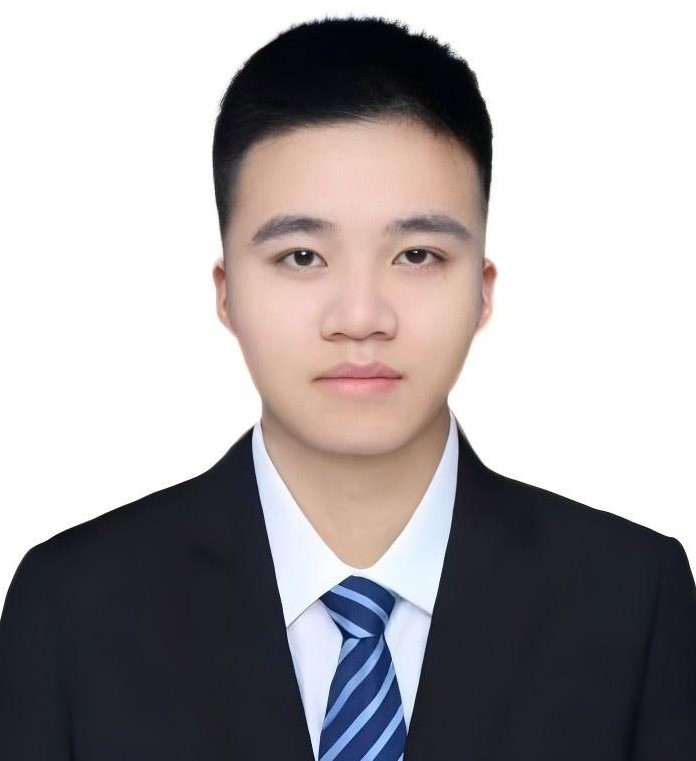}}]{QINGYANG ZENG} is currently pursuing a Master’s degree in Information Science and Technology at Beijing University of Chemical Technology, China. He received his Bachelor of engineering's degree from Beijing University of Chemical Technology in 2019. His current research interests are in natural language processing, focusing on multimodal entity recognition and large language models.
\end{IEEEbiography}
\vspace{-2.5in}  
\begin{IEEEbiography}[{\includegraphics[width=1in,height=1.25in,clip,keepaspectratio]{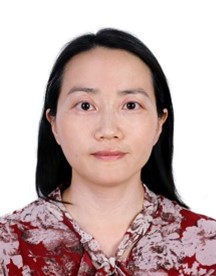}}]{JING WAN} received the B.S. degree in computer science from HUBEI University in 1997, the M.S. degrees in computer application from HUAQIAO University in 2000, and the Ph.D. degree in control theory and control engineering in 2011 from Beijing University of Chemical Technology. She is currently an Associate Professor at the Beijing University of Chemical Technology. Her research interests include knowledge graph and natural language processing.
\end{IEEEbiography}

\EOD

\end{document}